\documentclass[10pt,journal,cspaper,compsoc]{IEEEtran}

\usepackage{graphicx}
\usepackage{subfigure}
\usepackage{enumerate}
\usepackage{amsmath}
\usepackage{mathrsfs}
\usepackage{amssymb}
\usepackage{amsthm}
\usepackage{multirow}
\usepackage{rotating,makecell}
\usepackage{color}
\usepackage{algorithm}
\usepackage{algorithmic}
\usepackage[table]{xcolor}
\usepackage{colortbl}
\usepackage{subfloat}
\usepackage{url}
\usepackage{booktabs}
\usepackage{subfig}
\usepackage[table]{xcolor}

\graphicspath{
    {./figs/}
    }

\definecolor{lightgray}{gray}{0.9}
\definecolor{lightblue}{rgb}{0.98,0.98,1.0}

\newcommand{\lyj}[1]{\textcolor{black}{#1}}

\begin{document}

\makeatletter
\def\zxline#1{%
  \noalign{\ifnum0=`}\fi\hrule \@height #1 \futurelet
   \reserved@a\@xhline}

\title{\lyj{Video-based} Facial Micro-Expression Analysis: A Survey of \lyj{Datasets}, Features and Algorithms}%

\author{Xianye Ben,~\IEEEmembership{Member,~IEEE}, Yi Ren, Junping Zhang,~\IEEEmembership{Member,~IEEE}, Su-Jing Wang,~\IEEEmembership{Senior Member,~IEEE}, Kidiyo Kpalma, Weixiao Meng,~\IEEEmembership{Senior Member,~IEEE}, Yong-Jin Liu,~\IEEEmembership{Senior Member,~IEEE}
}

\markboth{To appear in IEEE Transactions on Pattern Analysis and Machine Intelligence, https://doi.org/10.1109/TPAMI.2021.3067464}
{Spontaneous Facial Micro-Expression Analysis}

\IEEEcompsoctitleabstractindextext{
\begin{abstract}
\lyj{Unlike} the conventional facial \lyj{expressions, micro-expressions are} involuntary and transient facial \lyj{expressions capable of revealing the} genuine \lyj{emotions} that people attempt to hide. Therefore, \lyj{they can provide important information in a broad range of} applications such as lie detection, criminal detection, etc. \lyj{Since micro-expressions are transient and of low intensity, however, their} detection and recognition \lyj{is} difficult and \lyj{relies} heavily on expert experiences. Due to its intrinsic particularity and complexity, \lyj{video-based} micro-expression analysis is attractive but challenging, and \lyj{has recently become} an active area of research. Although there \lyj{have been numerous} developments in this area, \lyj{thus far there has been no comprehensive survey that provides researchers with a systematic overview of these developments with a unified evaluation}. \lyj{Accordingly, in} this survey paper, we \lyj{first} highlight the key differences between macro- and micro-expressions, \lyj{then} use these differences to guide \lyj{our} research survey of \lyj{video-based} micro-expression analysis in a cascaded structure, \lyj{encompassing the} neuropsychological basis, datasets, features, \lyj{spotting} algorithms, recognition algorithms, applications and evaluation of \lyj{state-of-the-art approaches}. \lyj{For} each aspect, \lyj{the} basic techniques, advanced developments and major challenges are addressed and discussed. Furthermore, \lyj{after} considering the limitations \lyj{of} existing micro-expression datasets, we present and release a new dataset --- called \lyj{{\it micro-and-macro expression warehouse} (MMEW)} --- \lyj{containing} more video samples and more labeled emotion types. \lyj{We then perform a unified comparison of representative methods on CAS(ME)$^2$ for spotting, and on MMEW and SAMM for recognition, respectively.} Finally, some potential \lyj{future} research directions are explored and outlined.
\end{abstract}
\begin{IEEEkeywords}
Micro-expression \lyj{analysis}, \lyj{survey}, \lyj{spotting}, \lyj{recognition}, facial features, \lyj{datasets}
\end{IEEEkeywords}}

\maketitle
\IEEEdisplaynotcompsoctitleabstractindextext
\IEEEpeerreviewmaketitle

\section{Introduction}

\IEEEPARstart{E}{motions} \lyj{are an inherent part of} human life, and appear voluntarily or involuntarily through facial expressions when people communicate with each other face-to-face.
As a typical form of nonverbal communication, facial expressions play an important role in \lyj{the analysis of human} emotion \cite{1}, and have \lyj{thus} been widely studied in various domains (e.g., \cite{2,CorneanuSCG16}).

\begin{figure}[!t]
\centering
\includegraphics[width=0.54in]{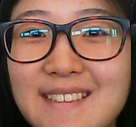}
\includegraphics[width=0.54in]{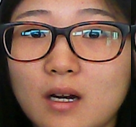}
\includegraphics[width=0.54in]{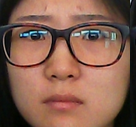}
\includegraphics[width=0.54in]{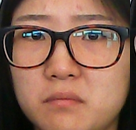}
\includegraphics[width=0.54in]{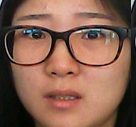}
\includegraphics[width=0.54in]{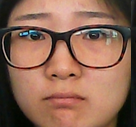}
\footnotesize
\makebox[0.6in]{Happiness}\makebox[0.6in]{Surprise}\makebox[0.58in]{Anger}\makebox[0.58in]{Disgust}\makebox[0.55in]{Fear}\makebox[0.6in]{Sadness}\\
\normalsize
\makebox[1.6in]{(a) Macro-expressions}\\\vspace{0.02in}
\includegraphics[width=1.1in]{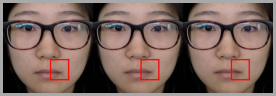}
\includegraphics[width=1.1in]{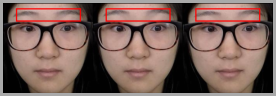}
\includegraphics[width=1.1in]{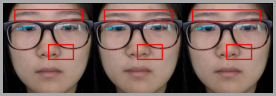}\vspace{-0.02in}
\footnotesize
\makebox[1.15in]{Happiness}\makebox[1.16in]{Surprise}\makebox[1.16in]{Disgust}\\\vspace{0.02in}
\includegraphics[width=1.1in]{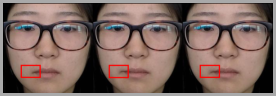}
\includegraphics[width=1.1in]{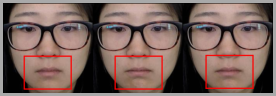}
\includegraphics[width=1.1in]{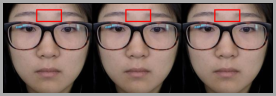}\vspace{-0.02in}
\footnotesize
\makebox[1.16in]{Fear}\makebox[1.16in]{Sadness}\makebox[1.16in]{Others}\\
\normalsize
\makebox[1.6in]{(b) Micro-expressions}\\
\vspace{-0.04in}
\caption{\lyj{Six macro-expressions and six micro-expressions, sampled from the same person, in the MMEW dataset (see Section \ref{sec:datasets}). While macro-expressions can be analyzed based on a single image, micro-expressions need to be analyzed across an image sequence due to their low intensity. For micro-expressions, the subtle changes outlined in red boxes are explained in the supplemental material.}}
\vspace{-0.2in}
\label{fig:macro-micro}
\end{figure}

Broadly speaking, \lyj{there are} two classes of facial expressions: macro- and micro-expressions \lyj{(Figure \ref{fig:macro-micro})}. The major difference between these two classes lies in both \lyj{their} duration and intensity. Macro-expressions are voluntary, usually last \lyj{for between} 0.5 \lyj{and} 4 seconds \cite{116}, and \lyj{are made using} underlying facial movements \lyj{that cover a} large facial area \cite{CorneanuSCG16}; \lyj{thus, they} can be clearly \lyj{distinguished} from noise such as eye blinks. \lyj{By} contrast, micro-expressions are involuntary, rapid and local expressions \cite{4,5}, \lyj{the} typical duration \lyj{of which is} between 0.065 and 0.5 seconds \cite{15}. Although people may intentionally conceal or restrain their real emotions by disguising \lyj{their} macro-expressions, micro-expressions are uncontrollable and can \lyj{thus} reveal the genuine emotions that humans try to conceal \cite{4,5,6,7}.

Due to \lyj{their inherent} properties \lyj{(}short duration, involuntariness and low intensity\lyj{)}, micro-expressions are very difficult to identify with the naked eye; \lyj{only} experts who have been \lyj{extensively trained} can distinguish micro-expressions. Nevertheless, even after intense training, \lyj{humans can recognize only 47\% of micro-expressions on average} \cite{10}.
\lyj{Moreover}, human analysis of micro-expressions is time-consuming, expensive and error-prone. Therefore, it is highly desirable to develop automatic systems for micro-expressions analysis  based on computer vision and pattern analysis techniques \cite{nag2019facial}.

Note that, in this paper, the concept of micro-expression analysis involves two aspects: \lyj{namely, spotting} and recognition.
\lyj{Micro-expression spotting involves identifying whether a given video clip contains a micro-expression, and if such an expression is found, identifying the onset (starting time), apex (the time with the highest intensity of expression) and offset (ending time) frames. Micro-expression recognition \lyj{involves classifying} a micro-expression into a set of predefined emotion types, e.g., happiness, surprise, sadness, disgust or anger, etc.}

\subsection{\lyj{Challenges and differences from conventional techniques}}

There have been a large number of various techniques proposed for conventional macro-expression analysis (see \lyj{e.g.} \cite{CorneanuSCG16} and reference therein). However, it is \lyj{non-trivial} to adapt these existing techniques to micro-expression analysis. Below, we list \lyj{the} three main technical challenges when compared to conventional macro-expression analysis.

{\it \lyj{(1) Challenges in collecting datasets}}. It is very difficult to elicit proper micro-expressions from participants in a controlled environment. \lyj{Moreover}, it is also difficult to correctly label these elicited micro-expressions as the ground truth, even for experts. Therefore, useful micro-expression datasets are scarce. \lyj{In the few datasets available, as summarized in Section \ref{sec:datasets}, the number of samples is much smaller than the case for conventional macro-expression datasets, which presents a key challenge for designing or training micro-expression analysis algorithms.}

{\it \lyj{(2) Challenges in designing spotting algorithms.}} \lyj{Macro-expression can usually be accomplished} using a single face image. However, due to \lyj{their} low intensity, it is almost impossible to detect micro-expressions in a single image; instead, an image sequence is \lyj{required}. Furthermore, \lyj{different from macro-expression detection,} spotting is a novel problem in micro-expression analysis. The techniques of micro-expression spotting \lyj{are summarized in Section \ref{sec:spotting}}.

{\it \lyj{(3) Challenges in designing recognition algorithms.}} \lyj{Like micro-expression spotting,} micro-expression \lyj{recognition} requires an image sequence. Usually, the image/video features used in macro-expression analysis can also be used in micro-expression analysis: \lyj{however,} special treatment \lyj{is} required \lyj{for the latter}. Due to \lyj{their} short duration and low intensity, the signals of these features are very weak, \lyj{while} indiscriminately amplifying these signals will amplify noises \lyj{from} head movement, illumination, optical flow estimation, etc. Section \ref{sec:features} summarizes \lyj{facial features for micro-expression} and Section \ref{sec:Recognition} summarizes the techniques of micro-expression recognition.

\subsection{\lyj{Contributions}}

\lyj{Over the past} decade, \lyj{research into} micro-expression analysis has been blossoming in terms of \lyj{datasets, spotting} and recognition techniques. However, the large \lyj{part} of micro-expression \lyj{research remains} scattered, and very few systematic surveys exist. \lyj{One} recent survey work \cite{102} focuses on the pipeline of a micro-expression recognition system from \lyj{an engineering perspective; a second survey} \cite{oh2018survey} provides a collection of results copied from the original papers. Different from \cite{102} and \cite{oh2018survey}, \lyj{this paper presents a comprehensive survey that makes the following contributions:
\begin{itemize}
\item Based on our summary of the limitations of existing micro-expression datasets, we present and release a new dataset, called {\it micro-and-macro expression warehouse} (MMEW), which contains more video samples and labeled emotion types than the existing datasets. Containing both macro and micro-expressions sampled from the same subjects, MMEW is the only one micro-expression dataset used for recognition for pretraining with macro-expression data from dataset itself instead of looking for other datasets. Researchers can also mine the relationship between the macro- and micro-expressions of the same subject for future research. In addition, the samples in MMEW have a larger image resolution (1920$\times$1080 pixels) than existing datasets.  This characteristic makes micro-expression clues display more incisively.
\item After the in-depth analysis of the differences between macro- and micro- expressions, we provide a comprehensive overview focusing on computing methodologies related to micro-expression spotting and recognition, as well as the common image and video features that can be used to build an appropriate automatic system. This survey also includes a detailed summary of up-to-date micro-expression datasets; based on this summary, we conduct subject-independent experiments which have the potential of being used in real-world micro-expression applications, and perform a fair comparison of representative micro-expression spotting and recognition methods on the MMEW and SAMM datasets.
\end{itemize}}

\lyj{The remainder of this paper} is organized as follows. Section \ref{sec:basis} carefully summarizes the difference between macro- and micro-expressions.
Section \ref{sec:datasets} summarizes existing datasets.
The image and video features useful for micro-expression analysis are collected and categorized in Section \ref{sec:features}. Representative algorithms are summarized in Sections \ref{sec:spotting} (for micro-expression spotting) and \ref{sec:Recognition} (for micro-expression recognition). Potential applications of micro-expression analysis \lyj{are} summarized in Section \ref{sec:applications}. Section \ref{sec:recommendation} presents a detailed comparison and recommendation of different methods. Some remaining challenges and future \lyj{research} directions are summarized in Section \ref{sec:direction}.  Finally Section \ref{sec:conclusion} presents our concluding remarks.

\section{Differences between macro- and micro- expressions}
\label{sec:basis}

\begin{figure}[tp]
\centering
\includegraphics[width=1.0\columnwidth]{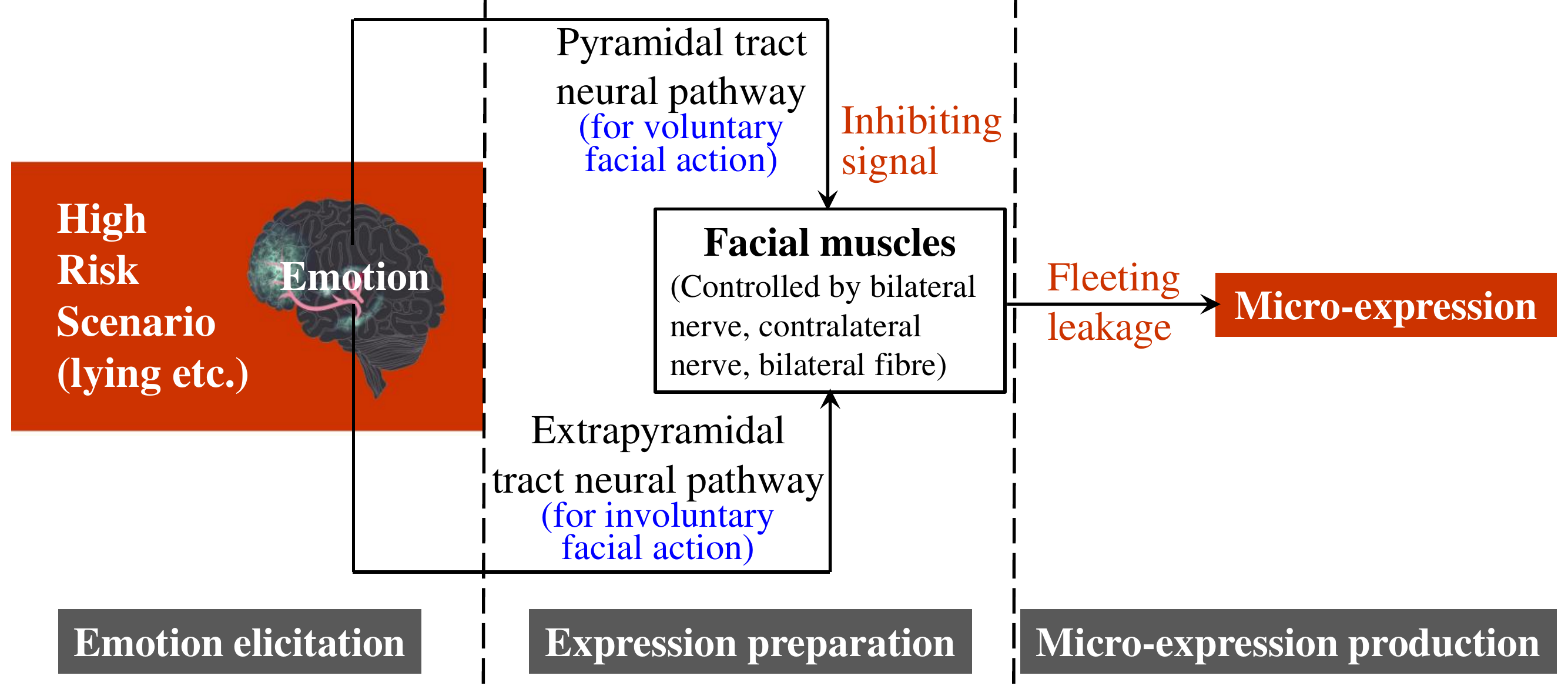}
\caption{\lyj{Two distinct neural pathways exist for conveying facial behavior: namely, the pyramidal and extrapyramidal tract neural pathways. The former is responsible for macro-expressions with voluntary facial actions, while the latter is responsible for spontaneous facial expressions. In high-risk scenarios, such as lying, both pathways are activated and engage in a back-and-forth battle, resulting in the fleeting leakage of genuine emotions in the form of micro-expressions.}
}
\vspace{-0.15in}
\label{fig:neuro-basis}
\end{figure}

Facial expressions are the results of \lyj{the} movement of facial skin and connective tissue. Facial muscles, which control these movements, are activated by facial nerve \lyj{nuclei}, which \lyj{are} in turn controlled by cortical and subcortical upper motor neuron circuits.
\lyj{One} neuropsychological study of facial expression \cite{Rinn1984} showed that \lyj{there are} two distinct neural pathways (located in different brain areas) for mediating facial behavior. The cortical circuit \lyj{is located} in the cortical motor strip and is primarily responsible for {\it posed} facial expressions (i.e., voluntary facial actions).
\lyj{Moreover,} the subcortical circuit, \lyj{which is located} in the subcortical areas of the brain, is primarily responsible for {\it spontaneous} facial expressions (i.e., involuntary emotion).
When people \lyj{attempt} to conceal or restrain their expressions in an \lyj{intensely} emotional \lyj{situation}, both systems are likely \lyj{to be} activated, resulting in the fleeting leakage of genuine emotions \lyj{in the form of} micro-expressions \cite{Matsumoto2011} \lyj{(Figure \ref{fig:neuro-basis})}. Throughout this paper, we \lyj{will focus on} spontaneous micro-expressions.

Micro-expressions are localized facial deformations caused by \lyj{the} involuntary contraction of facial muscles \cite{12}. \lyj{In comparison,} macro-expressions involve more muscle in a larger \lyj{facial} area and the intensity of muscle motion \lyj{is also} relatively stronger. \lyj{Compared} to macro-expressions, the intrinsic differences in \lyj{terms of neuroanatomical} mechanism is that micro-expressions have very short duration, slight variation and fewer action areas on the external facial features \cite{13,14}. This difference can also be concluded from the mechanism of concealment \cite{4}: when people try to conceal their feelings, \lyj{their true emotion can quickly `leak out'} and may be manifested as micro-expressions.

Other studies \lyj{have also shown} that the motion of muscles in the upper face (forehead and upper eyelid) during fake expressions are controlled by \lyj{the} precentral gyrus, which is dominated by \lyj{the} bilateral nerve, while the motion of muscle in the lower face is controlled by \lyj{the} contralateral nerve \cite{5,7,121}. The spontaneous expressions are controlled by the subcortical structure, the muscular movement of which is controlled by \lyj{the} bilateral fibre. This also provides evidence for the restraint hypothesis \cite{122}, \lyj{which states that} random expressions controlled by \lyj{the} pyramidal tract neural pathway \lyj{undergo} a back-and-forth battle with the spontaneous expressions controlled by the extrapyramidal tract; \lyj{during this process,} the extrapyramidal tract is dominated by bilateral fibers. Furthermore, if \lyj{activity in the} extrapyramidal tract is reflected on the face, it controls the muscles of \lyj{the} upper face, leading to spontaneous expressions \cite{5}. This also \lyj{leads to the} conclusion that when micro-expressions \lyj{occur}, they can happen \lyj{on} different parts of the face \cite{123}.

\lyj{The} hypothesis that feedback from \lyj{the} upper and lower facial regions has \lyj{different} effects on micro-expression recognition was proposed in \cite{zeng2018false}.
Three supporting studies were \lyj{subsequently} presented to highlight \lyj{the} three roles of facial feedback in judging the subtle movements of micro-expressions. First, when \lyj{feedback in the upper face} is enhanced by a restricting gel, micro-expressions with a duration of 450 ms can be more easily detected. Second, when \lyj{lower face} feedback is enhanced, the accuracy of sensing micro-expressions (duration conditions of  50, 150, 333, and 450 ms) is reduced. Third, blocking \lyj{lower face} feedback improves the recognition accuracy of micro-expressions.

{\it Characteristics of micro-expressions.}
Micro-expressions can express seven universal emotions: disgust, anger, fear, sadness, happiness, surprise and contempt \cite{18}. Ekman suggests that there are \lyj{certain facial muscles that cannot be consciously controlled, which he refers to as reliable muscles; these muscles} serve as sound indicators of the occurrence of related emotions \cite{125}. Studies have also shown that micro-expressions \lyj{may} contain all or only part of \lyj{the} muscle movements \lyj{that make up} common expressions \cite{125, 126,127}. Therefore, compared to macro-expressions, micro-expressions are facial expressions with short duration and \lyj{characterized by greater} facial muscle movement inhibition \cite{125, 126}, which can reflect \lyj{a person's} true emotions and \lyj{are more} difficult to control \cite{128}. Table~\ref{tab:macro-micro} summarizes the major differences between macro- and micro-expressions.

\makeatletter
\def\zxline#1{%
  \noalign{\ifnum0=`}\fi\hrule \@height #1 \futurelet
   \reserved@a\@xhline}

\begin{table}[t]
\centering
\caption{Main differences between macro- and micro-expressions.}
\renewcommand\arraystretch{1.3}
\begin{tabular}{l l l}
\zxline{1.3pt}
Difference & Micro-expression & Macro-expression \\
\zxline{1pt}
\rowcolor{lightgray} Noticeability & Easy to ignore & Easily noticed \\
\multirow{2}{*}{Time interval} & Short duration & Long duration \\
 & (0.065-0.5 seconds) & (0.5-4 seconds) \\
\rowcolor{lightgray} Motion intensity & Slight variation & Large variation \\
\multirow{2}{*}{Subjectivity} & Involuntary  & Voluntary \\
 & (uncontrollable) & (under control) \\
\rowcolor{lightgray} Action areas & Fewer &Almost all areas \\
\zxline{1.3pt}
\end{tabular}
\vspace{-0.15in}
\label{tab:macro-micro}
\end{table}

{\it Short duration.}
The short duration \lyj{is} considered to be the most important characteristic of micro-expressions. Most psychologists now agree that micro-expressions do not last \lyj{for} more than half a second. Yan et al. \cite{15} performed an elegant analysis that employed distribution curves of total duration and onset duration for the \lyj{purposes of} micro-expression analysis. \lyj{These authors} revealed that in a micro-expression, the start-up time (time from the onset frame to the apex frame) is usually within 0.26 seconds. Furthermore, the intensity of muscle motion in the micro-expression is very weak and \lyj{the expression itself is} uncontrollable due to the psychological inhibition of human instinctive response.

{\it Dynamic features.}
Neuropsychology shows that left and right \lyj{sides of the} face differ from each other in terms of expressing emotions \cite{5}: the left \lyj{side} expresses emotions that are more intense. Studies on facial asymmetry \lyj{have} also found that the right \lyj{side} expresses social context cues more conspicuously, while the left \lyj{side} expresses more personal feelings \cite{5}. These \lyj{studies provide further} evidence for the distinction between fake expressions and natural expressions, and thus indirectly explain the dynamic features of micro-expressions.

\section{Datasets of Micro-Expressions}
\label{sec:datasets}

The development of micro-expression analysis techniques \lyj{has been largely dependent} on well-established datasets with correctly labelled ground truth.
Due to \lyj{their} intrinsic characteristics, such as involuntariness, short duration and slight variation, eliciting micro-expressions in a controlled environment is very difficult.
\lyj{Thus far,} a few micro-expression datasets have been developed. However, as summarized in this section, most of \lyj{these} still have various deficiencies \lyj{as regards their} elicitation paradigms, labelling methods or small data size. Therefore, at the end of this section, we present and release a new dataset for micro-expression recognition.

\begin{table*}[t]
\caption{\lyj{Seven publicly released micro-expression datasets.}}
\vspace{-0.05in}
Here, {\it samples$^*$} refers to the samples of micro-expressions. Note that CAS(ME)$^2$ \lyj{and MMEW} also contain samples of macro-expressions.
Not all participants produce micro-expressions; here, {\it Participants$^*$} refers to the number of participants who generate micro-expressions. \lyj{The parentheses enclose the number of emotional video clips in that category.}
\vspace{-0.1in}
\begin{center}
\renewcommand{\arraystretch}{1.3}
\resizebox{\textwidth}{38mm}{
\begin{tabular}{l l lll l l l l l}
\zxline{1.3pt}
\multirow{3}*{Characteristics} & \multicolumn{9}{c}{Datasets}\\ \cline{2-10}
 & \multirow{2}*{MEVIEW}
 & \multicolumn{3}{c}{SMIC}
 & \multirow{2}*{CASME}
 &\multirow{2}*{CASME II}
 &\multirow{2}{*}{CAS(ME)$^2$}
 &\multirow{2}*{SAMM}
 &\multirow{2}*{MMEW}\\
 \cline{3-5}& &HS&VIS&NIR &  & & & & \\
\zxline{1pt}
\rowcolor{lightgray} Num of samples$^*$& \lyj{40} & 164 &71 & 71 &195 &247& 57 & 159 & 300 \\
Participants$^*$ & 16 &16 &8 &8  &35 &35 &22 &32 &36\\
\rowcolor{lightgray} Frame rate &25 &100 &25 & 25 & 60&200 &30 &200 &90\\
Mean age &N/A &N/A & & &22.03 &22.03 & 22.59 &33.24 &22.35 \\
\rowcolor{lightgray} Ethnicities &N/A &3 & & &1 &1 &1&13 &1\\
Resolution &1280$\times$720 &\multicolumn{3}{l}{640$\times$480} & 640$\times$480 \& 1280$\times$720 &640$\times$480 & 640$\times$480 &2040$\times$1088 &1920$\times$1080\\
\rowcolor{lightgray} Facial resolution &N/A &\multicolumn{3}{l}{190$\times$230} &150$\times$190 &280$\times$340& N/A & 400$\times$400 &400$\times$400\\
Emotion classes& \makecell[l]{\lyj{7} categories:\\ Happiness (6) \\Anger (2)\\ Disgust (1)\\ Surprise (9) \\Contempt (6)\\ Fear (3)\\ \lyj{Unclear (13)}} &\multicolumn{3}{c}{\makecell[l]{3 categories:\\Positive (107)\\Negative (116)\\ Surprise (83)} }   &\makecell[l]{8 categories:\\ Amusement (5) \\ Disgust (88)\\Sadness (6)\\Contempt (3)\\ Fear (2)\\ Tense (28)\\ Surprise (20) \\ Repression (40)} &\makecell[l]{5 categories:\\Happiness (33)\\ Repression (27)\\ Surprise (25)\\ Disgust (60)\\ Others (66)} &\makecell[c]{4 categories:\\ Positive (8) \\Negative (21)\\ Surprise (9)\\ Others (19)}
&\makecell[l]{\lyj{7 categories:}\\\lyj{Happiness (24)}\\ \lyj{Surprise (13)}\\ \lyj{Anger (20)}\\ \lyj{Disgust (8)}\\ \lyj{Sadness (3)}\\ \lyj{Fear (7)}\\ \lyj{Others (84)}}
&\makecell[l]{7 categories:\\Happiness (36)\\ Anger (8)\\ Surprise (89)\\ Disgust (72)\\ Fear (16)\\ Sadness (13)\\ Others (66)}\\
\rowcolor{lightgray}Available labels &Emotion/FACS &\multicolumn{3}{l}{Emotion}   &Emotion/FACS &Emotion/FACS &\makecell[l]{Emotion/FACS/\\Video type} &Emotion/FACS &Emotion/FACS\\
Download URL&\makecell[l]{http://cmp.fel\\k.cvut.cz/cechj\\/ME/ }&
\multicolumn{3}{c}{\makecell[l]{http://www.cs\\e.oulu.fi/SMIC\\Database} }& \makecell[l]{http://fu.psych.ac.cn\\/CASME/casme-en.\\php }&
\makecell[l]{http://fu.psych.\\ac.cn/CASME/\\casme2-en.php
}&
\makecell[l]{http://fu.psych.\\ac.cn/CASME/\\cas(me)2-en.php
}&
\makecell[l]{http://www2.\\docm.mmu.ac.\\uk/STAFF/M.\\Yap/dataset.php
}&
\makecell[l]{http://www.\\dpailab.com/\\database.html}
\\
\zxline{1.3pt}
\end{tabular}
}
\end{center}
\label{tab:datasetses}
\begin{flushleft}
\end{flushleft}
\vspace{-0.15in}
\end{table*}

\begin{figure*}[t]
\centering
\includegraphics[height=.0705\linewidth]{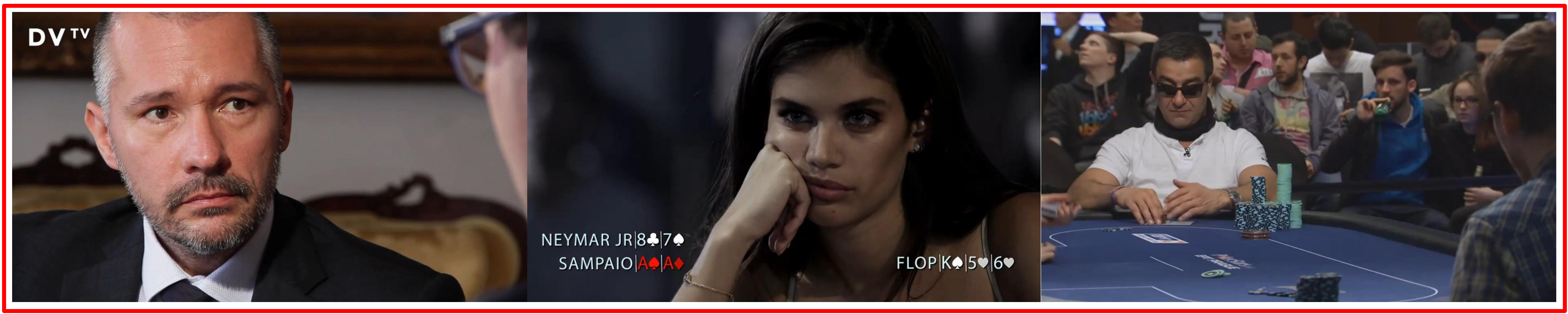}\hspace{2pt}
\includegraphics[height=.0705\linewidth]{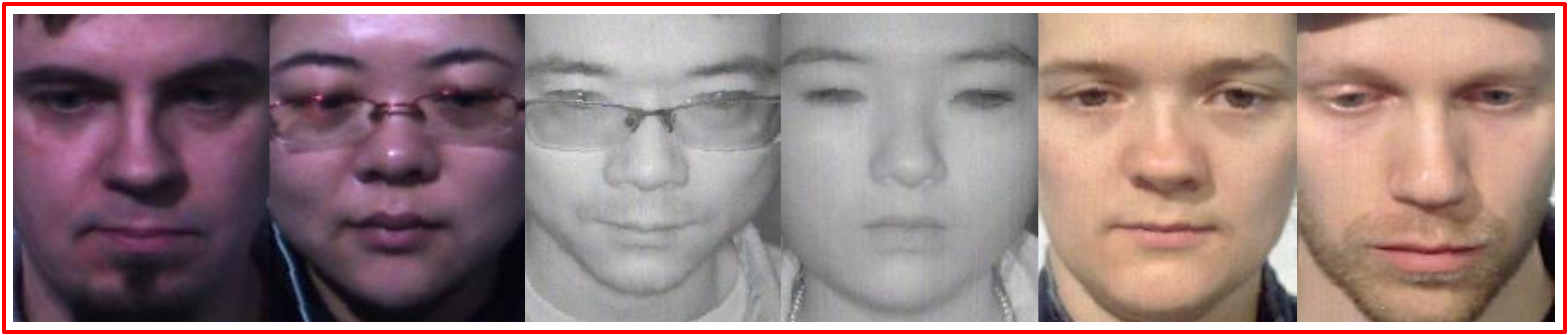}\hspace{2pt}
\includegraphics[height=.0705\linewidth]{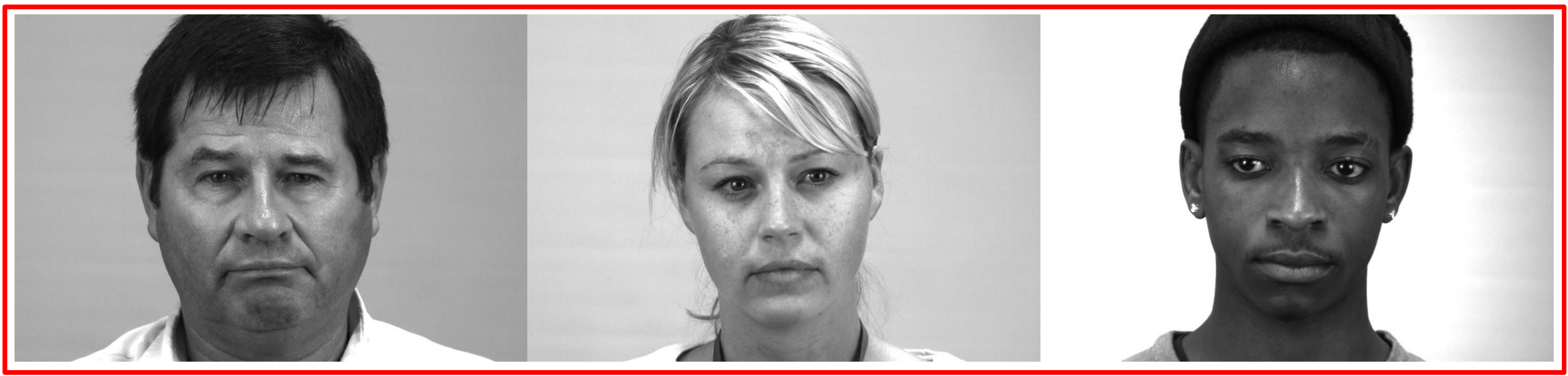}
\includegraphics[height=.069\linewidth]{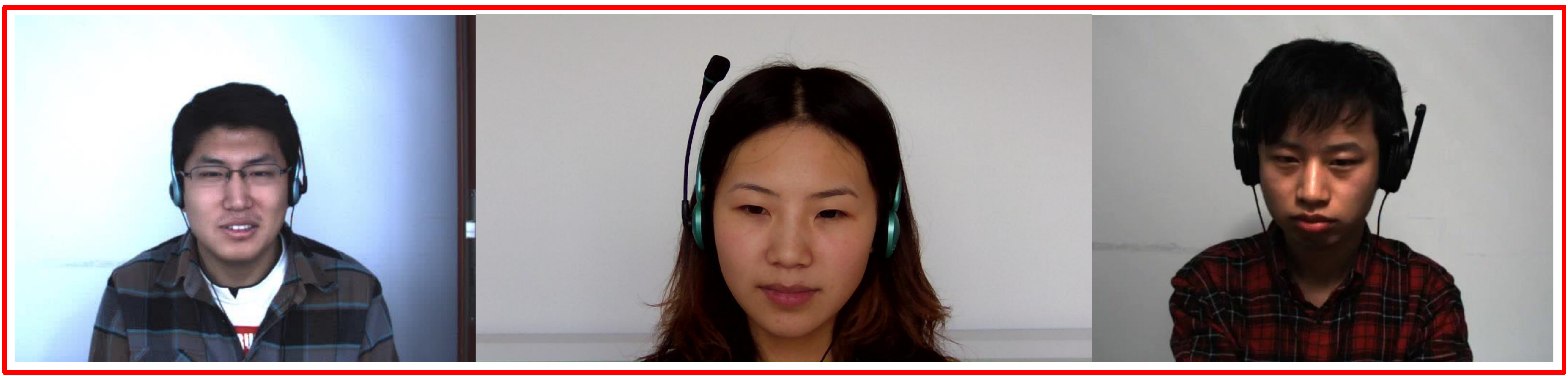}\hspace{2pt}
\includegraphics[height=.069\linewidth]{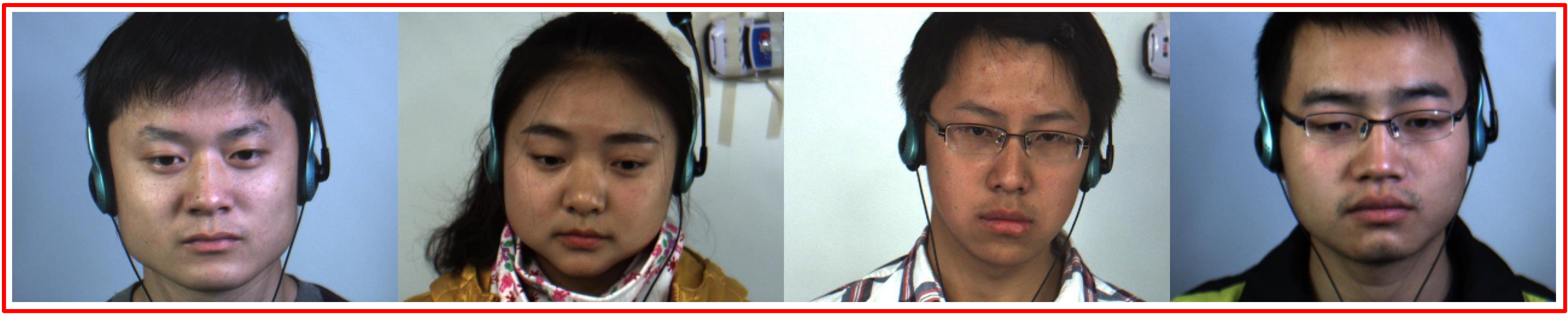}\hspace{2pt}
\includegraphics[height=.069\linewidth]{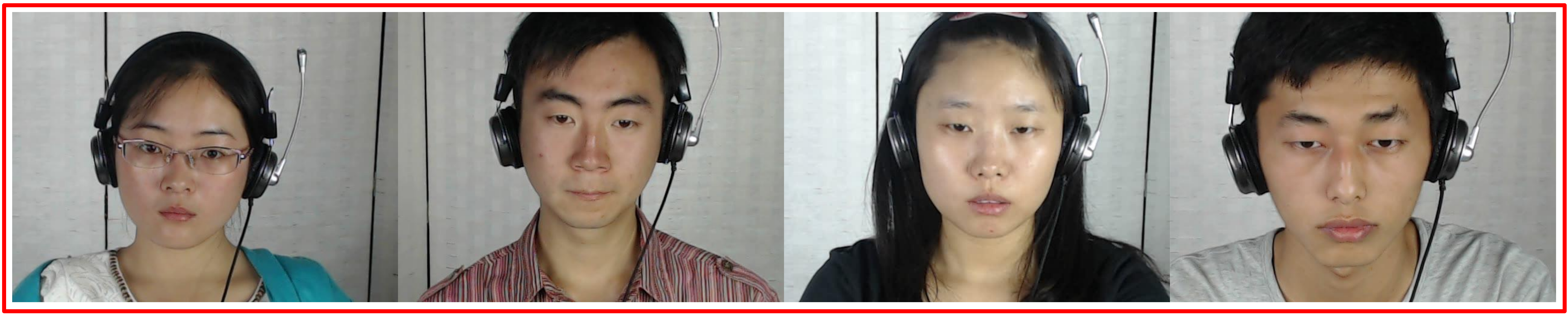}
\caption{Snapshots of six micro-expression datasets. From left to right and from top to bottom, \lyj{these are as follows:} MEVIEW \cite{110}, SAMM \cite{24}, SMIC \cite{21}, CASME \cite{22}, CASME II \cite{23}, and CAS(ME)$^2$ \cite{25}.}
\label{fig:dataset-snapshot}
\vspace{-0.15in}
\end{figure*}

Some early \lyj{studies} elicited micro-expressions by constructing high-stakes \lyj{situations; for example,} asking people to lie by concealing negative affect aroused by an unpleasant film and simulating pleasant feelings \cite{Ekman1974}, or \lyj{telling} lies about a mock theft in a crime scenario \cite{Frank1997}.
However, micro-expressions elicited in this way are often contaminated \lyj{by} other non-emotional facial movements, such as conversational behavior.

Since 2011, nine representative micro-expression datasets \lyj{have been} developed: namely, USF-HD \cite{32}, Polikovsky's dataset \cite{18}, York DDT \cite{19}, MEVIEW \cite{110}, SMIC \cite{21}, CASME \cite{22}, CASME II \cite{23}, SAMM \cite{24} and CAS(ME)$^2$ \cite{25}.
Both USF-HD and Polikovsky's dataset contain posed micro-expressions:
\begin{itemize}
\item in USF-HD, participants were asked to perform both macro- and micro-expressions, and
\item in Polikovsky's dataset, participants were asked to simulate micro-expression motion.
\end{itemize}
\lyj{It should be noted that these} posed micro-expressions are different from spontaneous ones.
\lyj{Moreover,} York DDT \lyj{is} made up of spontaneous micro-expressions with high ecological validity.
Nevertheless, similar to lie detection \cite{Ekman1974,Frank1997}, the data in York DDT was mixed with non-emotional facial movement \lyj{resulting from} talking.
Furthermore, all \lyj{of} three datasets (USF-HD, Polikovsky's dataset and York DDT) are not publicly available.

The \lyj{paradigm of elicitation} by telling lies has two major drawbacks:
\begin{itemize}
\item micro-expressions are often contaminated with irrelevant (i.e., talking) facial movements, and
\item \lyj{the} types of elicited micro-expressions are seriously restricted (e.g., the happiness type can never be elicited).
\end{itemize}
\lyj{By} contrast, it \lyj{is} well recognized that watching \lyj{an} emotional video \lyj{while maintaining a neutral expression (i.e.} suppressing emotions) is an effective method for eliciting micro-expressions.

Five \lyj{datasets} --- SMIC, CASME, CASME II, SAMM and CAS(ME)$^2$ --- used this elicitation paradigm.
MEVIEW used another quite different elicitation paradigm: \lyj{namely, constructing} a high-stakes situation by making use of poker games or TV interviews with difficult questions.

\lyj{In the below,} we first review MEVIEW, summarizing its advantages and disadvantages, and then \lyj{move on to focus} on the mainstream datasets (SMIC, CASME, CASME II, SAMM and CAS(ME)$^2$).
All \lyj{six of these} datasets, \lyj{which are} publicly available, \lyj{are} summarized in Table \ref{tab:datasetses}; \lyj{see} also Figure \ref{fig:dataset-snapshot} for some snapshots. For comparison, Table \ref{tab:datasetses} also includes our \lyj{newly} released dataset MMEW, which is presented \lyj{in more detail} at the end of this section.

MEVIEW \cite{110} consists of realistic video clips (i.e., shots in a non-lab controlled environment) from two scenarios: (1) some key moments in poker games, and (2) a person \lyj{being asked} a difficult question in a TV interview. Both scenarios \lyj{are notable for their} high stress factor. \lyj{For example}, in poker games, players \lyj{try} to conceal or fake their true emotions, and the key moments in \lyj{the} videos show the detail of a player's face while the cards \lyj{are} being uncovered: at these moments, micro-expressions are most likely to appear. \lyj{The} MEVIEW dataset contains \lyj{40} micro-expression video clips at 25 fps with a resolution \lyj{of} $1280\times 720$. The average length of \lyj{the} video clips in the dataset is 3 seconds and the camera shot is often switched. The emotion types in MEVIEW \lyj{are} divided into \lyj{seven} classes: happiness, contempt, disgust, surprise, fear, anger \lyj{and unclear emotions}.

The advantage of \lyj{the} MEVIEW dataset is that its scenarios are real, which will benefit the training or testing of micro-expression analysis algorithms. The disadvantages include \lyj{the following}: (1) in these real-scene videos, the participants were seldom shot from the frontal pose, \lyj{meaning that} the number of valid samples is quite small, and (2) the number of participants in the dataset is only 16, which is small.

\lyj{We will now} review SMIC, SAMM, CASME, CASME II and CAS(ME)$^2$. All the image sequences in these datasets are shots in controlled laboratory environments.

\lyj{The} SMIC dataset \cite{21} provides three data subsets with different types of recording cameras: HS (standing for high-speed camera), VIS (for normal visual camera) and NIR (for near-infrared camera). Since micro-expressions have \lyj{a} short duration and low intensity, a higher spatial and temporal resolution may help to capture more details. Therefore, the HS subset was collected with a high-speed camera \lyj{with a frame rate of} 100 fps. HS can be used to study the characteristics of \lyj{the} rapid changes of the micro-expressions. \lyj{Moreover,} the VIS and NIR subsets are used to add diversity of the dataset, \lyj{meaning that} different algorithms and comparisons \lyj{that address} more modes of micro-expressions can \lyj{possibly} be developed. Both \lyj{the} VIS and NIR subsets were collected with 25 fps and $640\times 480$ resolution. In contrast to a down-sampled version of the data with 100 fps, VIS can be used to study the normal behavior of micro-expressions, e.g. when motion blurs in web camera \lyj{footage} appear. NIR \lyj{images were photographed using} a near-infrared camera and can be used to eliminate the effect of lighting illumination on micro-expressions. The drawbacks of \lyj{the} SMIC dataset \lyj{pertain to} its labelling methods: (1) only three emotion labels (positive, negative and surprise) are provided, (2) the emotion labeling was only based on participants' self-\lyj{reporting, and} given that different participants may rank the same emotion \lyj{differently}, this overall self-\lyj{reporting} may not be precise, and (3) the labels of action units (AUs) are not provided in SMIC.

AUs are fundamental actions of individual muscles or groups of muscles, which are defined \lyj{according to} the Facial Action Coding System (FACS) \cite{16}, a system \lyj{used} to categorize human facial movements by their appearance on the face. AUs are widely used to characterize the physical expression of emotions. Although FACS has successfully \lyj{been used to code} emotion-related facial action in macro-expressions (Table \ref{tab:facs}), the AU labels in different emotion categories of micro-expressions are diverse and \lyj{consequently deserving of study in their} own right. Different from SMIC, AU labels were provided in CASME, CASME II, CAS(ME)$^2$ and SAMM.

\begin{table}[t]
\centering
\caption{Emotional facial action coding system for macro-expression}
\renewcommand\arraystretch{1.3}
\begin{tabular}{l l}
\zxline{1.3pt}
Emotion & Action units \\
\zxline{1pt}
\rowcolor{lightgray} Happiness & 6+12 \\
 Sadness & 1+4+15 \\
\rowcolor{lightgray} Surprise & 1+2+5B+26 \\
Fear & 1+2+4+5+7+20+26 \\
\rowcolor{lightgray}Anger& 4+5+7+23 \\
 Disgust & 9+15+16 \\
\rowcolor{lightgray}Contempt & R12A+R14A \\
\zxline{1.3pt}
\end{tabular}
\vspace{-0.15in}
\label{tab:facs}
\end{table}

CASME, CASME II and CAS(ME)$^2$ were developed by the same group \lyj{and utilized} the same experimental protocol. 
In a well-controlled laboratory environment, four lamps were chosen to provide steady and high-intensity illumination \lyj{(this approach can} also effectively avoid \lyj{the flickering of lights} caused by alternative current\lyj{)}. To elicit micro-expressions, participants were instructed to \lyj{maintain a neutral facial expression} when watching video episodes with high emotional valence.
CASME II is an extended version of CASME, \lyj{with the major differences between them being} as follows:
\begin{itemize}
\item CASME II used a high-speed camera \lyj{with a} sampling rate \lyj{of} 200 fps, while the sampling rate in CASME is only 60 fps;
\item CASME II has a larger face size (of $280\times 340$ pixels) in image sequences, while the face size in the CASME \lyj{samples} is only $150\times 190$ pixels;
\item CASME and CASME II have 195 and 247 micro-expression samples respectively. CASME II has \lyj{a more uniformly balanced number of samples across each emotion class}, i.e., Happiness (33 samples), Repression (27), Surprise (25), Disgust (60) and Others (66); \lyj{by contrast, the CASME samples are much more poorly distributed, with} some classes \lyj{having} very few samples (e.g., only 2 samples in fear and 3 samples in contempt; see Table \ref{tab:datasetses}).
\end{itemize}

\begin{figure}[t]
\centering
\vspace{-0.01in}
\includegraphics[width=1.0\columnwidth, trim= 0 70  0 0]{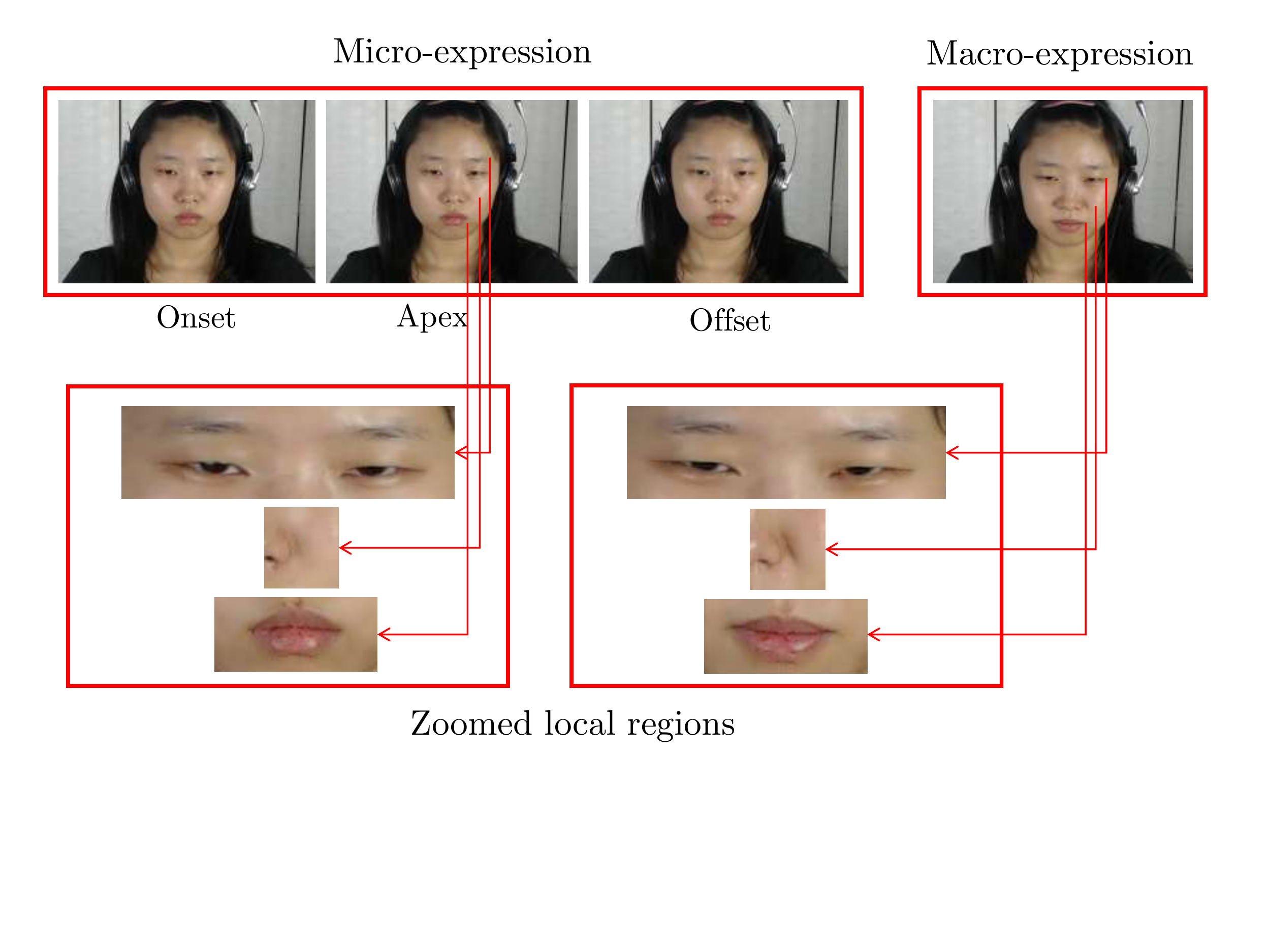}
\vspace{-0.4in}
\caption{Micro and macro-expression for the same person on the CAS(ME)$^2$ \lyj{dataset}. The first row is \lyj{the} original images, \lyj{while} the second row corresponds to the zoomed local regions of the micro-expression \lyj{apex} and the macro-expression of the same expression.}
\label{fig:schematic-micro_macro}
\end{figure}

Successfully eliciting micro-expressions by \lyj{maintaining neutral} faces is a difficult task \lyj{in itself; it} is even more difficult to elicit both macro- and micro-expressions.
CAS(ME)$^2$ collected this kind of data, in which macro- and micro-expressions are \lyj{distinguished} by duration time; i.e., whether or not it is smaller than 0.5 seconds. Figure \ref{fig:schematic-micro_macro} \lyj{presents} macro- and micro-expressions of the same participant from \lyj{the} CAS(ME)$^2$ dataset.
\lyj{The} CAS(ME)$^2$ dataset was divided into Part A and Part B: Part A contains 87 long videos with both macro and micro-expressions, \lyj{while} Part B contains 300 macro-expression samples and 57 micro-expression samples. The emotion classes includes positive, negative, surprise and others. All samples in CASME, CASME II and CAS(ME)$^2$ were coded \lyj{into} onset, apex and offset frames, and labeled based on a combination of AUs, the emotion types of emotion-evoking videos and self-reported emotion.

\lyj{The} SAMM dataset \cite{24} contains 159 samples (i.e., image sequences containing spontaneous micro-expressions) recorded by a high-speed camera with 200 fps and \lyj{a resolution of} $2040\times 1088$. Similar to the series of CASMEs, these samples were recorded in a well controlled laboratory environment with \lyj{carefully designed} lighting conditions, such that \lyj{light fluctuations} (which resulted in flickering on the recorded images) can be avoided. \lyj{A total of} 32 participants (16 males, 16 females, mean age 33.24) with a very good diversity of ethnicities (including 13 races) were recruited for the experiment. To assign emotion labels, in addition to self-\lyj{reporting}, each participant was required to fill in a questionnaire before starting the experiment, \lyj{after which} each emotional stimuli video \lyj{was} specially tailored for different participants to elicit \lyj{the} desired emotions in micro-expressions. Seven emotion categories were labelled in SAMM: contempt, disgust, fear, anger, sadness, happiness and surprise.

\begin{table}[!t]
\renewcommand{\arraystretch}{1.3}
\newcommand{\tabincell}[2]{\begin{tabular}{@{}#1@{}}#2\end{tabular}}
\caption{The micro-expression categories and AU labeling}
\label{tab:tab.2}
\vspace{-0.05in}
\centering
\renewcommand\arraystretch{1.3}
\begin{tabular}{l l l}
\zxline{1.3pt}
\tabincell{l}{Emotion \\categories}	&Main AU labeling type	&Quantity \\
\zxline{1pt}
\rowcolor{lightgray} Happiness	&AU6/AU12/AU6+AU12	&36 \\
Surprise	&AU5/AU26/AU1+AU2/AU18	&89 \\
\rowcolor{lightgray} Anger	&AU17+AU13/AU7/AU9	&8 \\
Disgust	&AU9/AU10/AU4+7/AU4+9	&72 \\
\rowcolor{lightgray} Fear	&AU4+5/AU20/AU4+11/AU14	&16 \\
Sadness	&AU14/AU17/AU14+17/AU1+2	&13 \\
\rowcolor{lightgray} Others	&AU4/AU18/AU4+14/AU38/AU4+7	&66 \\
\zxline{1.3pt}
\end{tabular}
\vspace{-0.15in}
\end{table}

So far, CAS(ME)$^2$ is \lyj{the} most appropriate for micro-expression spotting, since it is close to the real scenarios \lyj{in that it contains} both macro- and micro-expressions. However, it is not suitable for micro-expression recognition, \lyj{since} the number of micro-expression samples is small (only 57).
\lyj{Among SMIC, CASME, CASME II and SAMM, SAMM is the best for micro-expression recognition, but also has the limitation of small sample numbers. Accordingly, to both provide a suitable dataset for micro-expression recognition and inspire new research directions, we introduce a new dataset {\it micro-and-macro expression warehouse} (MMEW).}

MMEW follows the same elicitation paradigm \lyj{used in} SMIC, CASME, CASME II, SAMM and CAS(ME)$^2$, i.e., watching emotional video episodes while attempting to maintain a neutral expression. The full details of the MMEW dataset \lyj{construction process} are presented in the supplemental material.
\lyj{All} samples in MMEW \lyj{were} carefully calibrated \lyj{by experts} with onset, apex and offset frames, and the AU annotation from FACS \lyj{was used} to describe the facial muscle movement area (Table \ref{tab:tab.2}). Compared with \lyj{the} state-of-the-art CASME II, the major advantages of MMEW are as follows:
\begin{itemize}
\item The samples in MMEW have a larger image resolution ($1920\times1080$ pixels), while the resolution of samples in CASME II is $640\times480$ pixels. Furthermore, MMEW has a larger face size \lyj{in image sequences} of $400\times400$ pixels, while the face size in the CASME II \lyj{samples} is only $280\times340$ pixels.
\item MMEW and CASME II \lyj{contain} 300 and 247 micro-expression samples, respectively. MMEW has more elaborate emotion classes, i.e. Happiness (36), Anger (8), Surprise (89), Disgust (72), Fear (16), Sadness (13) and Others (66); \lyj{by contrast}, the emotion classes of Anger, Fear and Sadness are not included in CASME II. Furthermore, the class of Others in CASME II contains 102 samples (41.3\% \lyj{of} all samples).
\item 900 macro-expression samples with the same class category (Happiness, Anger, Surprise, Disgust, Fear, Sadness), acted \lyj{out} by the same group of participants, are \lyj{also} provided in MMEW (see Figure \ref{fig:macro-micro} for an example). \lyj{These} may be helpful for further cross-modal research (e.g., from macro- to micro-expressions).
\end{itemize}
\lyj{Compared with the state-of-the-art SAMM, MMEW contains more samples (300 vs. 159; see Table \ref{tab:datasetses}). Moreover, MMEW contains sufficient samples of both macro- and micro-expressions from the same subjects (Figure \ref{fig:macro-micro}). This new feature also opens up a promising new avenue of pretraining with macro-expression data from the own dataset, instead of looking for other datasets, which we summarize in Section \ref{ssec:evaluation-cross}}
MMEW is publicly available\footnote{\url{http://www.dpailab.com/database.html}}.

In Section \ref{sec:recommendation}, \lyj{we use CAS(ME)$^2$ to evaluate micro-expression spotting methods, and further use SAMM and MMEW to evaluate micro-expression recognition,} which can serve as baselines for the comparison of new methods developed in the future.

\section{Micro-Expression Features}
\label{sec:features}

Since \lyj{a} micro-expression is a subtle and indiscernible \lyj{motion of human facial muscles}, the effectiveness of \lyj{micro-expression spotting and recognition relies heavily} on discriminative features that can be extracted from image sequences. In this section, we review existing features \lyj{using} a hierarchical taxonomy.
Note that state-of-the-art deep learning techniques have been applied to micro-expression analysis in an end-to-end manner, so that hand-crafted features are less \lyj{necessary}. The related deep learning techniques will be reviewed in Section \ref{ssec:deep-learning}.

Throughout this section, all summarized features are computed on a \lyj{preprocessing} and normalized image sequence. The preprocessing \lyj{involves} detecting the facial region, identifying facial feature points, aligning \lyj{the} facial regions in each frame to remove head movements, normalizing or interpolating additional images \lyj{into} the original sequence, etc. The reader is referred to Sections 3.1 and 3.2 in \cite{102} for more details on preprocessing.

The classical voxel representation depicts a given image sequence in a spatiotemporal space $\mathbb{R}^3$, with three dimensions $x$, $y$ and $t$: \lyj{here,} $x$ and $y$ are pixel locations in one frame and $t$ is the \lyj{frame time}. At each voxel $v(x,y,t)$, gray or color values are assigned.
\lyj{Subsequently,} the local, tiny transient changes of facial regions \lyj{within} an image sequence can be effectively captured by local patterns of gray or color information at each voxel, which we \lyj{refer} to as a kind of {\it dynamic texture} (DT). DT \lyj{has been widely} studied in \lyj{the} computer vision field and is traditionally defined as image sequences of moving objects/scenes that exhibit certain stationarity properties in time (see e.g. \cite{DorettoCWS03ijcv}).
Here, we treat DT as an extension of texture --- local patterns of color at each image pixel --- from the spatial domain to the spatiotemporal domain in an image sequence.
There are three ways to make use of \lyj{this} local intensity pattern information:
\begin{itemize}
\item DT features in the original spatiotemporal domain $(x,y,t)\in\mathbb{R}^3$ (Section \ref{sssec:dt});
\item DT features in the frequency domain: by applying Fourier or wavelet transforms to a signal in $\mathbb{R}^3$,  the information in the spatiotemporal domain can also be dealt with \lyj{in the} frequency domain (Section \ref{sssec:frequency});
\item DT features in a transformed domain by tensor decomposition: by representing a color micro-expression image sequence as a 4th-order tensor, all the information in $(x,y,t)\in\mathbb{R}^3$ can be interpreted via tensor decomposition: \lyj{namely, the} facial spatial information \lyj{is} mode-1 and mode-2 of the tensor, temporal information is mode-3 and color information is mode-4 (Section \ref{sssec:tensor});
\item Optical flow features, which \lyj{indicate} the patterns of motion of objects (facial regions in our case) and are often computed by the changing intensity of pixels between two successive image frames over time, based on partial derivatives of the image signals (Section \ref{sssec:oflow}).
\end{itemize}

We \lyj{examine these four feature classes in more detail} in the following four subsections \lyj{and all presented features are summarized in Table S2 in the supplementary material}.


\subsection{DT features in \lyj{the} spatiotemporal domain}
\label{sssec:dt}

\subsubsection{LBP-based features}

Many DT features in the spatiotemporal domain are related to {\it local binary patterns from three orthogonal planes} (LBP-TOP) \cite{51}. The local binary pattern (LBP) \cite{52} on each plane is represented as
\begin{equation}
LBP_{P,r} = \sum_{p=0}^{P-1}s(g_{p}-g_{c})2^p,
\end{equation}
where $c$ is a pixel at frame $t$, referred \lyj{to} the center, $g_{c}$ is the gray value of $c$, $g_{p}$ denotes the gray value of the $p$th pixel, $P$ is the number of neighboring points located inside the circle of radius $r$ centered at $c$, and $s(x)$ is an indicator function
\begin{equation}
s(x)=\left\{
\begin{aligned}
1 & \quad \text{$x\ge0$} \\
0 & \quad \text{$x<0$}
\end{aligned}
\right.
\end{equation}
By concatenating \lyj{the} LBPs' co-occurrence statistics in three orthogonal planes, Zhao et al. \cite{51} proposed the novel LBP-TOP descriptor, \lyj{which} concatenates LBP histograms from three planes.
LBP-TOP has been successfully applied to \lyj{the recognition of} both macro- and micro-expressions \cite{21,51}.

Rotation-invariant features, such as LBP-TOP, do not fully consider all directional information. Ben et al. \cite{108} proposed hot wheel patterns (HWP), HWP-TOP, and dual-cross patterns from three orthogonal planes (DCP-TOP) based on DCP \cite{Ding}.
The experiments in \cite{108} show that these features, \lyj{when} enhanced by directional information description, can further improve the micro-expression recognition accuracy.

To \lyj{resolve the issue} that LBP-TOP may be coded repeatedly, Wang et al. \cite{57} proposed \lyj{an} LBP with six intersection points (LBP-SIP), which can delete the six intersections of repetitive coding. In doing so, it reduces redundancy and the histogram length, and \lyj{also} improves the speed. \lyj{Moreover, in order to preserve} the essential patterns \lyj{so as to improve} recognition accuracy and \lyj{reduce} redundancy, Wang et al. \cite{95} proposed the super-compact LBP-three mean orthogonal planes (MOP) for micro-expression recognition.  However, the accuracy of this approach is slightly worse when dealing with short video.  To make the features more compact and robust to \lyj{changes in} light intensity,  Huang et al. \cite{60} proposed a completed local quantization patterns (CLQP) \lyj{approach} that decomposes the local dominant pattern of the central pixel with the surrounding pixels into sign, magnitude and orientation respectively, \lyj{then} transforms it into a binary code. \lyj{They also} extended CLQP \lyj{into} 3D space, \lyj{referred to as} spatiotemporal CLQP (STCLQP) \cite{59}.

\subsubsection{Second-order features}

This class of features make use of second-order statistics to characterize new DT representations \lyj{of} micro-expressions \cite{niu2019discriminative}. Hong et al. \cite{79} proposed a second-order standardized moment average pooling (called 2Standmap) that calculates low-level features (such as RGB) and intermediate local descriptors (such as Histogram of Gradient Orientation; HIGO) for each pixel by means of second-order average and max operations.

John et al. \cite{84} proposed the re-parametrization of the second-order Gaussian jet for encoding LBP. \lyj{This approach} obtains a set of three-dimensional blocks; \lyj{based on these,} more robust and reliable histograms can be generated,  which are suitable for  different facial analysis tasks. Kamarol et al. \cite{88} proposed a spatiotemporal texture map (called STTM) to convolve an input video sequence with a 3D Gaussian kernel function in a linear space representation. By calculating the second-order moment matrix, the spatiotemporal texture and the histogram of the micro-expression sequence are obtained. This algorithm captures subtle spatial and temporal variance in facial expressions with lower computational complexity, and is \lyj{also} robust to illumination variations.

\subsubsection{Integral projection}

Integral projection --- which is a one-dimensional curved shape pattern --- have also been investigated in \lyj{the} micro-expression analysis \lyj{context}.
Integral projection can be represented by a vector in which each entry corresponds to a 1D position (\lyj{obtained} by projecting the image along a given direction) and the entry value is the sum of \lyj{the} gray values of \lyj{the} projected image pixels at this position.
Huang et al. \cite{82} proposed a spatiotemporal LBP with Integral Projection (STLBP-IP); \lyj{according to this approach,} LBP is evaluated on \lyj{the} integral projections of image pixels obtained in the horizontal and vertical directions.

The same research group \cite{82} also developed a revisited integral projection algorithm to maintain the shape property of micro-expressions, followed by LBP operators to further describe the appearance and motion \lyj{changes} from horizontal and vertical integral projections. They \lyj{then} proposed a discriminative spatiotemporal LBP with revisited integral projection (DiSTLBP-RIP) for micro-expression recognition \cite{103}. Finally, they used a new feature selection \lyj{method that} extracted discriminative information based on the Laplacian method.

\subsubsection{Other miscellaneous features}

In addition to the aforementioned features, Polikovsky et al. \cite{65} proposed a 3D gradient descriptor \lyj{that employs} AAM \cite{93} to divide a face into 12 regions. By calculating and quantizing the gradient in all directions of each pixel, then constructing \lyj{a} 3D gradient histogram in each region, this descriptor extends the plane gradient histogram to capture the correlation between frames. Another widely used feature is the histogram of oriented gradients (HOG) \cite{119} \lyj{incorporating} the convolution operation and weighted voting. \lyj{These authors} also proposed a histogram of image gradient orientation (HIGO) \cite{119}, \lyj{which} uses a simple vote \lyj{rather than} a weighted vote. HIGO can maintain good invariance of the geometric and optical deformation of the image. Moreover, the gradient or edge direction density distribution can better describe the local image appearance and shape, \lyj{while small} head motions can be ignored without affecting the recognition accuracy. \lyj{HIGO} is \lyj{particularly} suitable for occasions where the lighting conditions vary widely.

To maintain the invariance to \lyj{both geometric and optical deformations of images,} Chen et al. \cite{81} employed weighted features and weighted fuzzy classification to enhance the valid information \lyj{contained in} micro-expression sequences; \lyj{however, their recognition process is still expensive in terms of time cost.} Lu et al. \cite{91} proposed a Delaunay-based temporal coding model. This model divides the facial region into \lyj{smaller} triangular regions based on the detected feature points, \lyj{then} calculates the accumulated value of the difference between the pixel values of each triangular region in the adjacent frame \lyj{by means of} local temporal variations (LTVs). \lyj{This approach} encodes texture variations corresponding to \lyj{facial} muscle activities, \lyj{meaning} that \lyj{any} influence of personal appearance \lyj{that is} irrelevant to micro-expressions can be suppressed.

Wang et al. \cite{63} used robust PCA (RPCA) to decompose a micro-expression sequence into dynamic micro-expressions with subtle motion information. \lyj{More} specifically, they utilized an improved local spatiotemporal directional feature (LSTD) \cite{zhao2013visual} \lyj{in order to} obtain a set of directional codes for all six directions: i.e., XY(YX), XT(TX) and YT(TX). \lyj{Subsequently}, the decorrelated LSTD (DLSTD) was obtained by singular value decomposition (SVD) \lyj{in order to} remove the irrelevant information and thus \lyj{allow the} important micro-motion information \lyj{to be} emphasized. Zong et al. \cite{105} designed a hierarchical spatial division scheme for spatiotemporal descriptor extraction, and \lyj{further} proposed a kernelized group sparse learning (KGSL) model to process hierarchical scheme-based spatiotemporal descriptors. \lyj{This approach} can \lyj{effectively} choose a \lyj{good} division grid for different micro-expression samples, and is \lyj{thus} more effective for micro-expression recognition tasks.  Zheng et al. \cite{76} developed a novel multi-task mid-level feature learning algorithm \lyj{that} boosts the discrimination ability of low-level features extracted by learning a set of class-specific feature mappings.

\subsection{Frequency domain features}
\label{sssec:frequency}

The micro-expression sequence can be transformed into the frequency domain \lyj{by means of either} Fourier or wavelet transforms.
\lyj{This makes} relevant frequency features, such as amplitude and phase information, available for subsequent tasks.
For example, the local geometric features \lyj{(such as the corners of facial contours and the facial lines, which are easily overlooked by some feature description algorithms)} can be easily identified by high-frequency information.

Oh et al. \cite{61} extracted the magnitude, phase and orientation of the transformed image \lyj{by means of} Riesz wavelet transform.
To discover the intrinsic two-dimensional local structures (i2D) of micro-expressions, Oh et al. \cite{77} also performed \lyj{a} Fourier transform to restore the phase and orientation of the i2D \lyj{via} a high-order Riesz transformation \lyj{using a} Laplacian of Poisson (LOP) band-pass filter; \lyj{this is} followed by extracting LBP-TOP features and the corresponding feature histogram after quantification. An advantage \lyj{of this approach} is that the i2D can extract some easy-to-lose local structures, such as complex facial contours. Similar to i2D, the i1D \cite{78} is extracted \lyj{through the use of a} first-order Riesz transformation. To obtain directional statistical structures from micro-expressions, Zhang et al. \cite{89} used the Gabor filter to obtain \lyj{the} texture images of important frequencies and suppress other texture images.

Different from \lyj{the} aforementioned research, Li et al. \cite{49} utilized Eulerian video magnification (EVM) to magnify the subtle motion in a video. \lyj{In more detail}, the representation of the frequency domain of the micro-expression sequence was obtained \lyj{via} Laplace transform, and some frequency bands were band-fed
to enhance the corresponding scale of the movement and achieve directional amplification of the micro-expressions. \lyj{Refining the EVM approach}, Oh et al. \cite{75} proposed \lyj{the} Eulerian motion magnification (EMM) method, \lyj{which} consists of amplitude-based EMM (A-EMM) and phase-based EMM (P-EMM). LBP-TOP was \lyj{then} used to extract the features of the micro-expression sequence, such that \lyj{their algorithm enables the micro-movements to be magnified}.

\subsection{DT features in tensor-decomposition spaces}
\label{sssec:tensor}

Treating the micro-expression sequence as a tensor \lyj{enables} rich structural spatiotemporal information \lyj{to be extracted} from it. Due to \lyj{their} high dimensionality, many tensor-based dimension reduction algorithms --- which \lyj{keep} the inter-class distances as large as possible and the intra-class distances as small as possible --- can be used \lyj{to obtain} a more effective discriminant subspace.

\lyj{By} viewing a gray-valued micro-expression sequence as a three-dimensional spatiotemporal tensor, Wang et al. \cite{85} proposed a discriminant tensor subspace analysis (DTSA) that preserves some useful spatial structure information. \lyj{More specifically}, this method projects the micro-expression tensor to a low-dimensional tensor space \lyj{in which} the inter-class distance is maximized and intra-class distance is minimized.

Ben et al. \cite{33} proposed a maximum margin projection with tensor representation (MMPTR) \lyj{approach. This method also views} a micro-expression sequence as a third-order tensor, \lyj{and} can directly extract discriminative and geometry-preserving features \lyj{by} maximizing the inter-class Laplacian scatter and minimizing the intra-class Laplacian scatter.

To obtain better discriminant performance, Wang et al. \cite{80} proposed tensor independent color space (TICS). Representing a micro-expression sequence in three RGB channels, \lyj{this approach} extracted features from the four-order tensor space by utilizing LBP-TOP to estimate four projection matrices, each \lyj{representing} one side of the tensor data.

\subsection{Optical flow features}
\label{sssec:oflow}

Optical flow is a motion pattern of moving objects/scenes in an image sequence, which can be detected by the intensity change of pixels between two image frames over time.
Many elegant optical flow algorithms \lyj{have been proposed} that are suitable for diverse application scenarios \cite{98}.
In micro-expression research, optical flow has been investigated as an important feature.

\begin{table*}[t]
\caption{Comparison of representative micro-expression spotting algorithms}
\vspace{-0.05in}
\begin{center}
\footnotesize
\renewcommand\arraystretch{1.2}
\vspace*{-7pt}
\begin{tabular}{>{\columncolor{lightgray}}p{2.5em} p{4em} p{18em} p{15em} p{16em}}
\zxline{1.3pt}
 \rowcolor{white} &References	&Detection/spotting results	&Advantages	&Disadvantages \\
 \zxline{1pt}
\rowcolor{lightgray}&\cite{32} \cite{36}&\makecell[l]{Distinguishing macro-expression\\ from micro-expression}	&&\\
\rowcolor{lightgray}&\cite{37}	&Micro-expression detection/spotting &	& \\
\rowcolor{lightgray}&\cite{46} \cite{GUO2021} &Onset, apex and offset frames detection	&  \multirow{-4}*{\makecell[l]{Able to capture tiny expression\\ variations}} & \multirow{-4}*{\makecell[l]{A threshold is manually set, \\when the training data is small.}}  \\
\multirow{2}*{\rothead{\makecell[l]{Optical flow\\ based methods}}}
&\cite{111}	& \makecell[l]{Spotting facial movements from\\ long-term videos}	& \makecell[l]{Able to obtain more accurate\\ features of movement} & Time-consuming\\
\zxline{1pt}
\rowcolor{lightgray}&\cite{18} \cite{39}	&\makecell[l]{Onset, apex and offset frames \\detection} &Simple &\makecell[l]{Only suitable for posed micro-\\expression, not spontaneous\\micro-expression}\\
&\cite{40} \cite{119}&Micro-expression spotting&Satisfactory results&\makecell[l]{Very complicated; parameters and\\ thresholds are manually set}\\
\rowcolor{lightgray}&\cite{43}	&Micro-expression spotting &\makecell[l]{Error due to head motion is \\minimized} &  \\
\rowcolor{lightgray}&\cite{49}	&\makecell[l]{Onset, apex and offset frames\\ detection}&\makecell[l]{Combination of two different \\features is more powerful} & \multirow{-3}*{\makecell[l]{A threshold is not easy to determine}} \\
\multirow{1}*{\rothead{\makecell[l]{Feature-descriptor-based \\methods}}}&\cite{41} \cite{48}	& Apex frame detection& \makecell[l]{Parameters and thresholds are set \\automatically}& Only able to spot the apex frame\\
\zxline{1.3pt}
\end{tabular}
\end{center}
\label{tab:spotting}
\vspace{-0.15in}
\end{table*}

Patel et al. proposed \lyj{the} spatiotemporal integration of optical flow vectors (STIOF) \cite{46}, which computes optical flow vectors inside small local spatial regions and \lyj{then} integrates these vectors into the local spatiotemporal volumes. Liong et al. proposed an optical strain weighted features (OSWF) algorithm \cite{71}, \lyj{which} extracts the optical strain magnitude for each pixel and \lyj{uses the feature extractor to form} the final feature histogram. Xu et al. \cite{97} proposed facial dynamics map (FDM) to capture small facial motions based on optical flow estimation \cite{98}.

Wang et al. \cite{111} proposed a main directional maximal difference (MDMD) \lyj{algorithm} to characterize the magnitude \lyj{of} maximal difference in the main direction of optical flow features.
More recently, Liu et al. proposed the Mean Directional Mean Optical Flow (MDMO) feature, which integrates the magnitude and direction of the main optical flow vectors from
a total of 36 non-overlapping regions of interest (ROIs) in a human face \cite{99}. \lyj{While} MDMO is a simple and effective feature, the average MDMO \lyj{operation} often loses the underlying manifold structure inherent in the feature space. \lyj{To address this}, the same research group further proposed a sparse MDMO \cite{124} by constructing a dictionary containing all the atomic optical flow features in the \lyj{entire} video, \lyj{as well as} applying a time pool \lyj{to achieve} the sparse representation. MDMO and sparse MDMO are similar to the histogram of oriented optical flow (HOOF) feature \cite{67}; \lyj{however, the key} difference is that MDMO and sparse MDMO use HOOF features in a local way (i.e., in each ROI region) and are \lyj{thus} more discriminative.

\subsection{\lyj{Deep features}}

\lyj{In addition to the aforementioned hand-crafted features, deep learning methods that can automatically extract optimal deep features have also been applied recently to micro-expression analysis; we summarize this class of methods in Section \ref{ssec:deep-learning}. In these deep network models, the feature maps preceding the full connected layers (particularly the last full connected layer) can usually be regarded as deep features. However, these deep models are often designed as black boxes, and the interpretability or explainability of these deep features is frequently poor \cite{Ras2018}.}

\subsection{Feature interpretability}

Compared with deep features which are implicitly learned through deep network optimization based on big data, conventional hand-crafted features are usually designed based on either human experiences or statistical properties, which have higher explainability. For examples, LBP-based features explicitly reflect a statistical distribution of local binary patterns. Integral projection method based on difference images can preserve the shape attributes of facial images. Optical flow feature-based method is a normalized statistic feature that considers both local statistic motion information and its spatial location.

\section{Spotting algorithms}
\label{sec:spotting}

\lyj{In the literature, micro-expression detection and spotting are two related but often confused terminologies. In our study, we define micro-expression detection as the process of identifying whether or not a given video clip (or an image sequence) contains a micro-expression. Moreover, we put emphasis on micro-expression spotting, which goes beyond detection: in addition to detecting the existence of micro-expressions, spotting also identifies three time spots, i.e. onset, apex and offset frames, in the whole image sequence:
\begin{itemize}
\item the onset is the first frame at which a micro-expression starts (i.e., changing from the baseline, which is usually the neutral facial expression);
\item the apex is the frame at which the highest intensity of the facial expression is reached;
\item the offset is the last frame at which a micro-expression ends (i.e., returning back to the neutral facial expression).
\end{itemize}}

\lyj{In the early stages} of micro-expression research, \lyj{manual} spotting based on the FACS coding system \lyj{was used} \cite{16}.
\lyj{However, manual} coding is laborious and time-consuming; e.g., recording a one-minute micro-expression video sample takes two hours \lyj{on average} \cite{29}.
Moreover, \lyj{manual} coding is subjective due to \lyj{differences in the} cognitive ability and living background of the participants \cite{30};
it is \lyj{therefore} highly desirable to develop automatic algorithms for spotting micro-expressions.

\lyj{There are three major challenges associated with} developing accurate and efficient micro-expression spotting algorithms.
First, detecting micro-expressions usually \lyj{relies} on setting the optimal upper and lower thresholds for \lyj{any} given feature \lyj(see Section \ref{sec:features}:)
the upper threshold aims at distinguishing micro-expressions from macro-expressions, \lyj{while} the lower threshold defines the minimal motion amplitude of micro-expressions.
Second, different people may \lyj{perform} different extra facial actions. For example, some people blink habitually, while other people \lyj{sniff more frequently}, which can cause movement in the facial area. The impact of these facial motion areas on \lyj{expression spotting should thus} be taken into consideration.
Third, when recording videos, many comprehensive factors \lyj{(including} head movement, physical activity, recording environment, and lighting) may significantly influence the micro-expression spotting.

\lyj{Existing} automatic micro-expression spotting algorithms can be broadly divided into two classes: \lyj{namely,} optical-flow-based and feature-descriptor-based methods. For \lyj{ease of} reading, we highlight the pros and cons of these two classes in Table~\ref{tab:spotting}.

\subsection{Optical flow-based spotting}

Optical flow algorithms can be used to measure the intensity change of image pixels over the time. Shreve et al. \cite{32,36} \lyj{proposed a strain pattern that was used as a measure of motion intensity} to detect micro-expressions. However, this method relies on manually selecting thresholds to distinguish \lyj{between} macro-expression and micro-expression. Furthermore, this method was designed \lyj{to detect} posed micro-expressions, but not spontaneous micro-expressions (\lyj{note} that posed and spontaneous micro-expressions vary widely in terms of facial movement intensity, muscle movement, and time intervals). Shreve et al. \cite{37} used optical flow to exploit non-rigid facial motion by capturing the optical strains. This method can achieve an 80\% true positive rate with a 0.3\% false positive rate in spotting micro-expressions. Moreover, this method can \lyj{also} plot the strains and visualize a micro-expression \lyj{as it occurs} over the time.

\begin{table*}
\footnotesize
\caption{Micro-expression recognition algorithms}
\vspace{-0.05in}
\centering
\renewcommand{\arraystretch}{1.1}
\begin{tabular}{>{\columncolor{lightgray}}p{3.1em} p{14em} p{18.5em} p{23em}}
   \zxline{1.3pt}
	\rowcolor{white} &\makecell[l]{Algorithm} &	Advantages	&Disadvantages \\
\zxline{1pt}
\rowcolor{lightgray} &\makecell[l]{SVM  \cite{80} \cite{97} \cite{99} }& Most commonly used &Restricted by the limited training samples\\
&\makecell[l]{ELM \cite{58} \cite{63} \cite{85}}&Obtains an optimal solution&Restricted by the limited training samples\\
\rowcolor{lightgray}\multirow{2}*{\rothead{\makecell[l]{ Traditional\\classifiers}}}&KNN \cite{67} \cite{69}&Simple to use &\makecell[l]{Classification performance needs to be further\\ improved}\\
\zxline{1pt}
&Selective CNN \cite{100}	 &Avoids irrelevant deep features&The fitness function is subjective to overfitting\\
 \rowcolor{lightgray}&DTSCNN \cite{112}&Avoids the overfitting problem & Highly dependent on hardware   \\
 &DWLD +DBN \cite{113} &\makecell[l]{Reduces the unnecessary learning for the \\redundant features } &  Insufficient samples  \\
\rowcolor{lightgray} &TLCNN \cite{114} &\makecell[l]{Solves the problem of limited \\micro-expression samples} &  Complicated   \\
\multirow{-2}*{\rothead{\makecell[l]{Deep learning}}}
&\makecell[l]{ELRCN \cite{115}}&\makecell[l]{Solves the problem of limited \\micro-expression samples }&\makecell[l]{There are shortcomings in preprocessing \\techniques and data augmentation}\\
\zxline{1pt}
\rowcolor{lightgray}	&\makecell[l]{DR	\cite{104}}&\makecell[l]{Handles the unsupervised cross-dataset\\ micro-expression recognition problem	} &\makecell[l]{The consistent feature distribution between the\\ training and testing samples is seriously broken }\\
&\makecell[l]{Source domain targetized \cite{109}}	&\makecell[l]{Utilizes efficient speech data to enha-\\nce micro-expression recognition accuracy}	
& \makecell[l]{Large difference of data distribution}  \\
\rowcolor{lightgray} 	
&SVD \cite{107}	&&  \\
\rowcolor{lightgray}&Coupled metric learning \cite{108}&\multirow{-2}*{\makecell[l]{Solves the limited samples problem}}&\multirow{-2}*{\makecell[l]{Based on an assumption that a linear transformation \\exists between macro- and micro-expressions}} \\
\multirow{-1}*{\rothead{\makecell[l]{Transfer learning\\from other domains}}}
&\makecell[l]{Auxiliary set selection model + \\a transductive transfer regression \\model \cite{zong2019toward}}	&\makecell[l]{Selects a small number of representative \\samples from the target domain data, \\rather than choosing all of them}	
& \makecell[l]{Difficult to select the best sample number}  \\\zxline{1.3pt}
\end{tabular}
\label{tab:recognition}
\vspace{-0.15in}
\end{table*}

Patel et al. \cite{46} used \lyj{a} discriminative response map fitting (DRMF) model \cite{47} to locate key points of the face based on \lyj{the FACS system. This method then} groups the motion vectors of key points (indicated by the optical flow) and computes a cumulative value of the motion amplitude, \lyj{shifted over time, to detect the onset, apex and offset frames} of a micro-expression. However, this method also \lyj{requires} a manually specified threshold. Wang et al. \cite{111} used the magnitude maximal difference in the main direction of \lyj{the} optical flow features to spot micro-expressions. Guo et al. \cite{GUO2021} proposed a magnitude and angle combined optical flow feature for micro-expression spotting, and obtained more accurate results than the method \lyj{proposed in} \cite{111}.

\subsection{Feature descriptor based spotting}

Polikovsky et al. \cite{18,39} proposed to use the gradient histogram descriptor and K-Means algorithm to locate the onset, apex and offset frames of posed micro-expressions. \lyj{While} their method is simple, \lyj{it only} works for posed micro-expressions \lyj{(not for spontaneous micro-expressions)}. Moilanen et al. \cite{40} divided the face into 36 regions, calculated the LBP histogram of each region, \lyj{then} used the Chi-square distance between the current frame and the average to \lyj{determine} the degree of change in the video. \lyj{While} this method is novel, the design concept is \lyj{somewhat} complicated and the parameters in this method also need to be set manually.

Davison et al. \cite{119} aligned and cropped faces for each video. This method \lyj{involves splitting} these faces into blocks and \lyj{calculating} the HOG for each frame. Afterwards, this method used Chi-Squared distance to calculate \lyj{the} dissimilarity between frames at a set interval \lyj{in order to spot} micro-expressions. Xia et al. \cite{43} modeled the geometric deformation \lyj{using an} active shape model with SIFT descriptor \cite{44} to locate the key points. \lyj{Their method then} performed the Procrustes transform between each frame and the first frame to remove the bias \lyj{caused by} the head motion. \lyj{Subsequently}, this method evaluated an absolute feature and its relative feature and combined them to estimate the transition probability. Finally, the micro-expression was detected according to a threshold. This method can minimize the error caused by head movement, but an optimal threshold is difficult to determine.

Li et al. \cite{49} made use of \lyj{the} Kanade-Lucas-Tomasi algorithm \cite{50} to track three specific points of every frame. This method extracted LBP and HOOF features from each area and obtained the differential value of every frame, then detected the onset, apex and offset frames based on a threshold. This method \lyj{successfully fuses} two different features to get more discriminative information; \lyj{however, again,} the threshold is difficult to determine.

Yan et al. \cite{41} and Han et al. \cite{HanLLL18icip} proposed to \lyj{using} feature difference \lyj{to locate} the apex frame of a micro-expression. This method \cite{41} used a constrained local model (CLM) \cite{42} to locate 66 key points in a face, and then divided the face into subareas relative to these key points. \lyj{Subsequently}, they calculated the correlation between \lyj{the} LBP histogram feature vectors of every frame and first frame. Finally, the frame with the highest correlation value was regarded as the apex frame of a micro-expression.
\lyj{Different from the above methods,} Liong et al. \cite{48} spotted the apex frame by utilizing \lyj{a} binary search strategy and restricting the region of interest (ROI) to a predefined facial sub-region; \lyj{here, the ROI selection} was based on the landmark coordinates of the face. Meanwhile, three distinct feature descriptors, \lyj{namely,} CLM, LBP and Optical Strain (OS), were adopted to further confirm the reliability of the proposed method. \lyj{Although} the methods in \cite{41} and \cite{48} did not require \lyj{the} manual setting of parameters and thresholds, \lyj{they are only able to} spot the apex frame. Some disadvantages of \lyj{the above-mentioned feature-descriptor-based} methods include \lyj{their} high computational cost and \lyj{the} instability caused by noise and illumination variation.

\section{Recognition algorithms}
\label{sec:Recognition}

Micro-expression recognition usually \lyj{comprises} two steps: feature extraction and feature classification.
Traditional classification methods rely on artificially designed features (e.g., those summarized in Section \ref{sec:features}).
State-of-the-art deep learning methods can automatically infer an optimal feature representation and offer an end-to-end classification solution.
\lyj{However,} deep learning requires a large number of training samples, while all existing micro-expression datasets are small;
\lyj{therefore,} transfer learning that makes use of the knowledge from a related domain (in which large datasets are available) has also been considered in micro-expression recognition.
In this section, we summarize these three classes of recognition methods: traditional methods, deep learning and transfer learning methods.
For \lyj{ease of} reading, we highlight the pros and cons of these methods in Table \ref{tab:recognition}.

\subsection{Traditional classification algorithms}

Pattern classification/recognition has a long history \cite{Duda2000PC}, and many well-developed classifiers \lyj{have already been proposed}; the reader is referred to \cite{Delgado2014WNH} for more details. In the special application of micro-expression recognition, many classic classifiers have been applied, \lyj{including the} support vector machine (SVM) \cite{80,97,99}, extreme learning machine (ELM) \cite{58,63,85} and $K$ nearest neighbor (KNN) classifier \cite{67,69}, to \lyj{name only} a few.

\subsection{Deep learning}
\label{ssec:deep-learning}

In recent years, deep learning \lyj{approaches that integrate} automatic feature extraction and classification in an end-to-end manner \lyj{have been} great success. These deep models can obtain state-of-the-art predictive performance in many applications including micro-expression recognition \cite{van2019capsulenet,verma2019learnet,xia2019spatiotemporal}.

Hao et al. \cite{113} proposed an efficient deep network for micro-expression recognition. This network \lyj{involves} a two-stage strategy: the first stage \lyj{use a} double Weber local descriptor for extracting initial local texture features, and the second stage \lyj{uses} a deep belief net (DWLD+DBN) for global feature \lyj{extraction}.
Wang et al. \cite{114} proposed a transferring long-term convolutional neural network (TLCNN), which used Deep CNN to extract features from each frame of micro-expression video clips.
Khor et al. \cite{115} proposed an Enriched Long-term Recurrent Convolutional Network (ELRCN) \lyj{that operates} by encoding each micro-expression frame into a feature vector \lyj{via} CNN. \lyj{It then} predicts the micro-expression by passing the feature vector through a Long Short-term Memory (LSTM) \lyj{module}. Through the enrichment of the spatial dimension (\lyj{via} input channel stacking) and the temporal dimension (\lyj{via} deep feature stacking), TLCNN and LSTM can effectively deal with the problem of \lyj{small sample size}.

To tackle the overfitting problem, Patel et al. \cite{100} developed a selective deep model that can remove irrelevant deep information from micro-expression recognition. Their model has a good generalization ability. Peng et al. \cite{112} \lyj{further} proposed a dual temporal scale convolutional neural network (DTSCNN) for micro-expressions recognition. DTSCNN used different streams \lyj{to adapt} to different frame rates of micro-expression video clips, each of which included an independent shallow network to \lyj{prevent overfitting}. These shallow networks were fed with optical-flow sequences to ensure that higher-level features \lyj{could} be extracted.

Since the deep network models have a huge number of parameters and weights, sufficiently well-labelled micro-expression samples are urgently needed \lyj{in these models} to improve the recognition performance.

\subsection{Transfer learning from other domains}

Due to the small data size \lyj{of} existing micro-expression datasets, transfer learning \lyj{--- which} transfers knowledge from a source domain (in which a large sample size is available) to a target domain \lyj{---} has been considered in micro-expression recognition. Usually, the source and target domains are different but related (e.g., macro- and micro-expressions) so that they share some common knowledge (e.g., macro- and micro-expressions have similar AUs and DT information when expressing emotions) and \lyj{benefitting from transferring this knowledge} the performance in the target domain \lyj{can be improved}.

Zong et al. \cite{104} proposed an effective framework, called domain regeneration (DR), for cross-dataset micro-expression recognition. The training and testing samples \lyj{are derived} from different micro-expression datasets. The DR framework is able to learn a regenerator that regenerates samples with similar feature \lyj{distributions} in source and target micro-expression datasets.

\lyj{Another} research group developed a series of transfer learning works \cite{108,109,107} on micro-expression recognition. In \cite{107}, they proposed a macro-to-micro transformation model using singular value decomposition (SVD). This model takes advantage of sufficiently labelled macro-expression samples to increase the \lyj{number} of training samples. In \cite{108}, \lyj{the authors} extracted several local binary operators (that jointly characterize macro- and micro-expressions) and transferred these operators into a common subspace shared by source and target domains for learning \lyj{purposes}. In \cite{109}, they proposed to transfer the knowledge of speech samples in the source domain to the micro-expression samples in the target domain, \lyj{in such a way that} the transferred samples \lyj{would have a} similar feature distribution in the target domain. All these three transfer learning methods have been shown to achieve better recognition accuracy than some previous works.

\lyj{Moreover, rather than} transferring learning from macro-expressions or speech samples, a recent work \cite{zong2019toward} proposed an auxiliary set selection model (ASSM) and a transductive transfer regression model (TTRM) to bridge the gap between the source and target micro-expression domains. This method outperforms many state-of-the-art approaches, \lyj{as the} feature distribution difference between the source and target domains \lyj{is small}.

\section{Applications}
\label{sec:applications}

Micro-expression analysis \lyj{has a wide range of potential} applications \lyj{in the fields of} criminal justice, business negotiation and psychological consultation, etc. \lyj{In the below,} we present the details of lie detection, which serves as a common and typical application scenario for micro-expression analysis.


\lyj{One} automatic lie detection instrument \lyj{that is widely used today is} the polygraph. \lyj{This instrument usually} collects multi-track physiology information such as respiration, pulse, blood pressure, pupil, skin electricity, and brain waves. If the \lyj{test subject} tells a lie, this instrument will \lyj{record} a fluctuation in some of the above physiological signals. \lyj{However,} whether the polygraph will work properly depends on many factors, such as the external environment during the test, the intelligence level of the tested person, the physical and mental condition of the tested person, and even the \lyj{skill level} of the polygraph operator. Sometimes, due to fear \lyj{experienced as a result of the unfamiliar polygraph testing environment}, some innocent people might show a state of panic and anxiety during the test, leading to false \lyj{positives in the test results}. On the other hand, some well-trained professionals may \lyj{be able to pass a polygraph even while lying}. In such a situation, micro-expression analysis can provide another \lyj{method of} lie detection.

Ekman \cite{6} pointed out that as an effective clue \lyj{for use when identifying} lies, micro-expressions \lyj{have} broad applications in lie detection. By detecting the occurrence of micro-expressions and understanding the emotion type behind \lyj{them}, it will be possible to accurately identify the true intentions of the participant and improve the rate of success in lie detection.
For example, when the participant reveals a micro-expression of happiness, this may indicate \lyj{hidden delight at having successfully passed the test} \cite{116}. However, when the participant shows the micro-expression of surprise, this may indicate that the participant has never considered or understood \lyj{the} relevant questions. Because the micro-expression is difficult to hide and is usually the same as \lyj{a person's} true state of mind when lying, psychologists believe that \lyj{the} micro-expression is an important clue for lie detection. P{\'e}rez-Rosas et al. \cite{perez2015deception} discovered that the five micro-expressions most related to falsehood are: frowning, eyebrow raising, lip corners \lyj{turning} up, lips protruded and head \lyj{turning to the side}.

\section{\lyj{Comparison}}
\label{sec:recommendation}

The research on micro-expression features, spotting and recognition algorithms, such as \lyj{that} summarized in this paper, are scattered \lyj{throughout the} literature. A great challenge is that their performance \lyj{results} are reported \lyj{across} different experimental settings, \lyj{making it difficult to conduct} a fair comparison based on a common framework. In this section, we present a study \lyj{that compares} a selected set of representative spotting and recognition algorithms. As pointed out in Section \ref{sec:datasets}, \lyj{thus far,} the CAS(ME)$^2$ dataset is most appropriate for spotting evaluation \lyj{(Section \ref{ssec:evaluation-spotting}), where the SAMM and MMEW datasets are most suitable for recognition evaluation (Section \ref{ssec:evaluation-recog}). By taking the advantage of the fact that it contains both macro- and micro-expressions of the same subjects, we also compared the experimental results by using different datasets for pre-training, and tested the performance on the MMEW and SAMM datasets (Section \ref{ssec:evaluation-cross}).} By ensuring that all the factors (including data sample size and pre-processing) are the same, the analysis provided in this section can serve as a baseline for the evaluation of new algorithms \lyj{to be designed} in future work.

\subsection{Evaluation of spotting algorithms}
\label{ssec:evaluation-spotting}

\begin{figure}[t]
\centering
  \includegraphics[width=0.9\columnwidth]{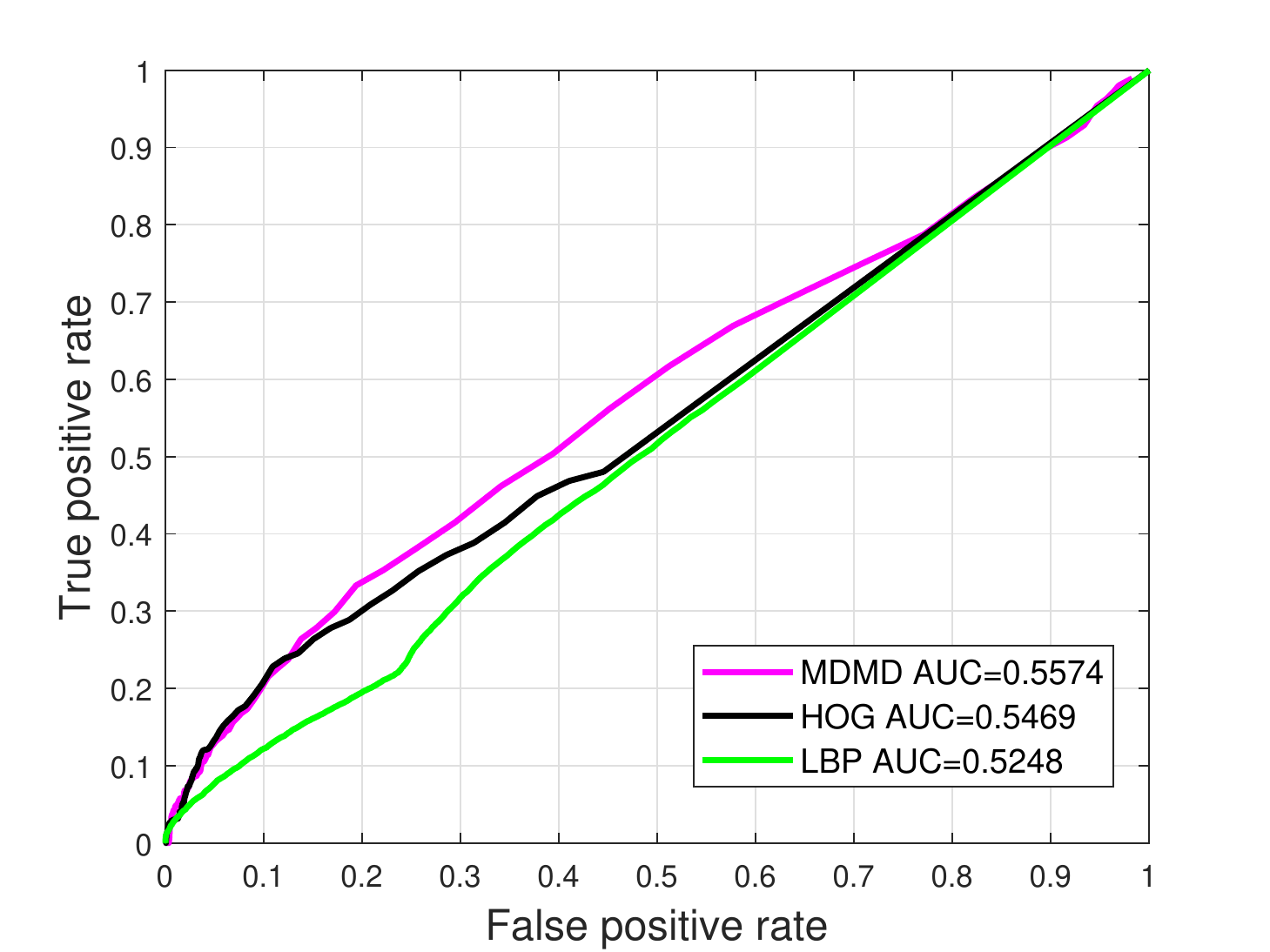}
    \caption{\lyj{ROC results of three methods --- MDMD, HOG and LBP --- on CAS(ME)$^2$ in terms of micro-expression spotting.}}
\label{fig:roc}
\vspace{-0.15in}
\end{figure}

First, we study the influence of the feature extraction methods \lyj{on} the performance of micro-expression spotting. \lyj{We use} the MDMD \cite{111}, HOG \cite{119} and LBP \cite{40} methods \lyj{for comparison in our spotting experiments} on CAS(ME)$^2$. For MDMD and  LBP, the micro-expression samples are divided into 5$\times$5 and 6$\times$6 blocks, respectively. For HOG, the number of blocks is 6$\times$6, the signed gradient direction binning is set to 2$\pi$, and the number of direction bins is set to 8. Figure \ref{fig:roc} \lyj{plots} the ROC curves of these three methods on CAS(ME)$^2$; the results show that MDMD performs the best in spotting micro-expressions. The possible reason is that MDMD applies the magnitude maximal difference in the main direction of optical flow features to detect/spot micro-expressions, leading to more notable differences or discriminant ability than those from \lyj{the} HOG or LBP features.

\subsection{Evaluation of recognition algorithms}
\label{ssec:evaluation-recog}

Compared to spotting, micro-expression recognition strongly depends on the amplification of subtle features, and then usually \lyj{requires} an additional preprocessing step. In Section \ref{sssec:preprocess}, we first evaluate the effects of different preprocessing methods \lyj{on the MMEW dataset}. \lyj{Then we perform a unified comparison on \lyj{the} MMEW and SAMM datasets to} evaluate the recognition accuracies of traditional methods (using hand crafted features) in Section \ref{sssec:traditional-methods} and state-of-the-art methods (including deep learning methods) in Section \ref{sssec:deep-learning}, respectively.

Throughout this section, all recognition results were obtained \lyj{under} the following settings. In the MMEW dataset, 234 samples from 6 classes (i.e., happiness, surprise, anger, disgust, fear, sadness) were used\footnote{Because transfer learning methods were included in our comparison, we exclude 66 samples in the ``Others'' category.}; in the SAMM dataset, 72 samples from 5 classes (i.e., happiness, surprise, anger, disgust, fear) were used\footnote{\lyj{We exclude 3 samples from the ``Sadness'' category (due to its small sample size) and 84 samples from the ``Others'' category due to the inclusion of transfer learning in our comparison.}}. In both datasets, all samples were randomly split into five subsets according to  ``subject independent'' approach, and the number of subjects in each subset is equal. Therefore,  this random split approach ensures that there is no overlap subject between the test  and the training sets at the same time. Then, five-fold cross-validation was performed, after which the average recognition results were reported.

\subsubsection{Evaluation \lyj{of different preprocessing methods}}
\label{sssec:preprocess}

Image sequence alignment and interpolation are two common preprocessing methods \lyj{utilized} in micro-expression recognition. We analyze the influences of these two methods by keeping the other modules unchanged. The details are summarized \lyj{below}.


{\bf Alignment algorithms.}
In the micro-expression recognition task, it is necessary for the micro-expression sequence to normalize the size of faces and align the face shapes \lyj{across all of} the different video samples. After all background parts are removed, only the facial areas in each video are preserved. \lyj{More specifically,} each image is normalized \lyj{to a size of} 231 $\times$ 231 pixels. Since the largest number of frames among all the image sequences \lyj{is 108},  the micro-expression image sequences are interpolated to the maximal value of 110 frames.
Here, we evaluate three alignment algorithms, \lyj{namely} ASM+LWM \cite{23}, DRMF+Optical flow alignment (OFA) \cite{99} and joint cascade face detection and alignment (JCFDA) \cite{28}, for face alignment on the MMEW dataset.  The LBP-TOP is used as a baseline to evaluate these three alignment algorithms.

\begin{figure}[t]
\centering
\includegraphics[height=.2\linewidth]{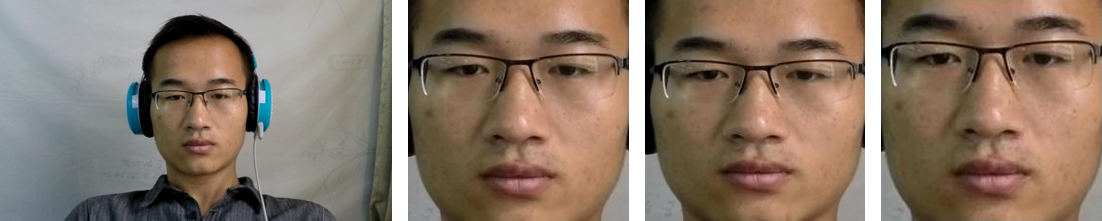}
\caption{Comparisons of three alignment algorithms. Due to the \lyj{space limitations}, we only show four frames \lyj{here}. From left to right: (a) an original image sequence; (b) results yielded by ASM+LWM \cite{23}; (c) results yielded by DRMF+OFA \cite{99}; (d) results yielded by JCFDA \cite{28}.}
\vspace{-0.15in}
\label{fig:three}
\end{figure}

\begin{figure}[t]
\centering
\includegraphics[width=1\linewidth]{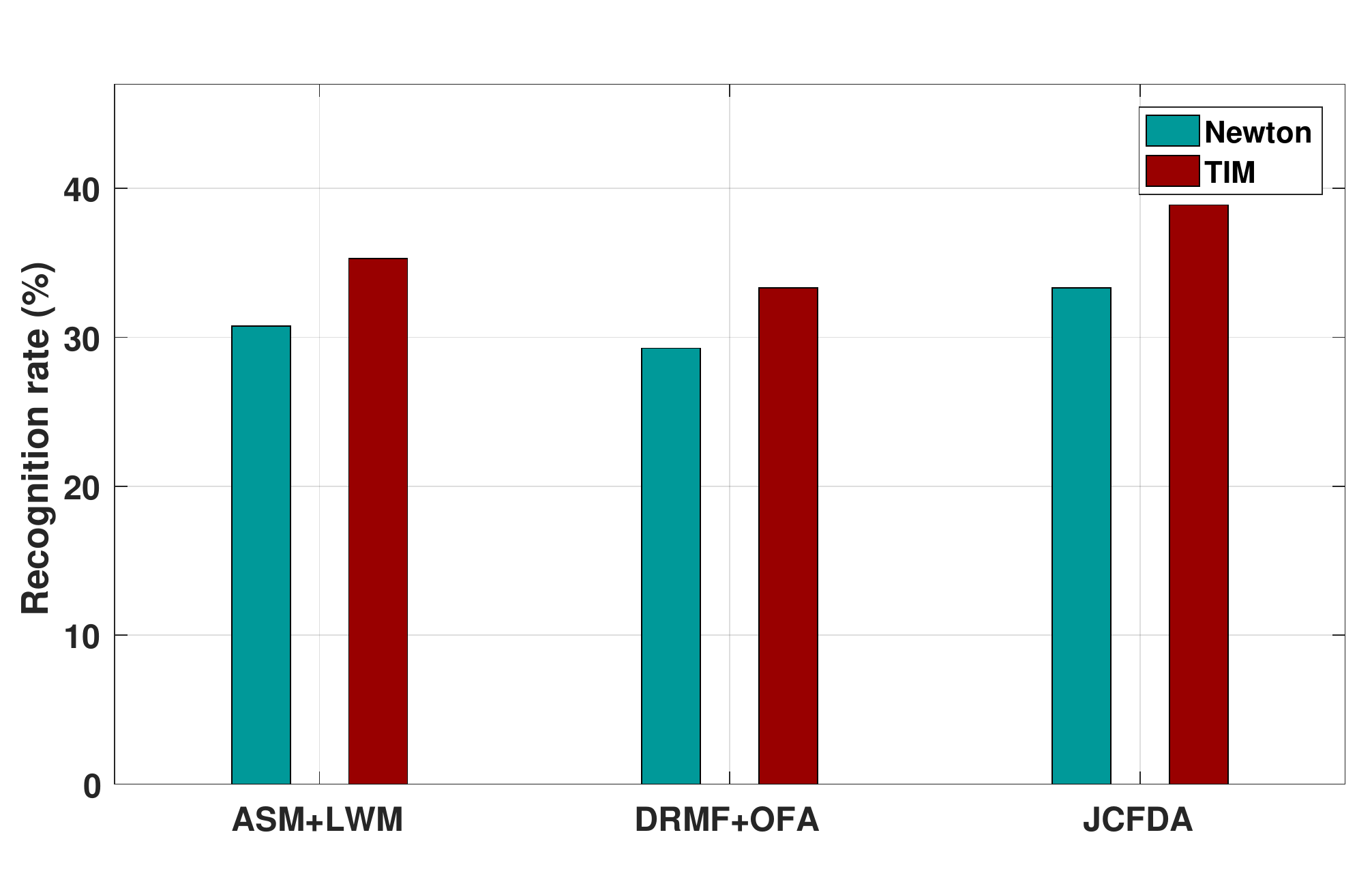}
\caption{The highest recognition rates of \lyj{LBP-TOP with three alignment and two interpolation algorithms on MMEW.}}
\label{fig:three1}
\vspace{-0.15in}
\end{figure}

In the settings of LBP-TOP, the radii $R_{x}$, $R_{y}$ of axes $X$ and $Y$ vary from 1 to 4. In order to avoid having too many parameter combinations, $R_{x}$ is set to be equal to $R_{y}$, and the variation range of the radius $R_{t}$ of \lyj{the} $T$ axis is also ranged from 1 to 4. Thus, the numbers of neighborhood points of \lyj{the} $XY$, $XT$ and $YT$ planes are all set to be 8. The recognition rates of these three alignment algorithms are obtained based on the uniform pattern under different parameters and different radius configurations. Finally, an SVM classifier with RBF kernel \cite{118} is applied. The experimental results are presented in Figures \ref{fig:three}, \ref{fig:three1} \lyj{and S3 in supplemental material. From the figures, we can} draw the following conclusions.
\lyj{First, compared with the other} two methods, JCFDA \cite{28} provides the best recognition rate of 38.9\%.  The reason is that JCFDA \cite{28} combines face detection with alignment and learns at the same time under a cascade framework; \lyj{this} joint learning greatly improves the alignment and real-time performance. In the following experiments, we thus choose JCFDA \cite{28} to preprocess the micro-expression image sequences.

%

{\bf Interpolation algorithms.}
Two interpolation algorithms, \lyj{namely} Newton interpolation \cite{33} and TIM interpolation \cite{20}, are used to interpolate micro-expression sequences into different numbers of frames, including 10, 20, 30, 40, 50, 60, 70, 80, 90, 100, 110 frames.
The highest recognition rates of these two algorithms with different interpolated frame numbers are \lyj{provided in Figure S4 in supplemental material}. We \lyj{can} observe that as the number of frames increases, the recognition rates of \lyj{the} two interpolation algorithms increase \lyj{initially} and then decrease \lyj{towards} the end. Newton interpolation can obtain its highest recognition rate of  33.3\% under the interpolated frame number of 60. Meanwhile, the recognition rate of TIM is relatively higher compared with that of the Newton interpolation, \lyj{with} the highest recognition rate \lyj{reaching} 38.9\% when the micro-expression sequence is respectively interpolated to 30 or 60 frames. Accordingly, in the following experiments, TIM is used to interpolate each micro-expression sequence to 60 frames.


\begin{table}[t]
\footnotesize
\caption{\lyj{Recognition rates (\%) of micro-expressions using hand-crafted features with different classifiers on MMEW and SAMM.}}
\centering
\renewcommand{\arraystretch}{1.2}

\begin{tabular}{p{9.3em} p{1.9em}<{\centering} p{1.9em}<{\centering} p{1.9em}<{\centering} p{1.9em}<{\centering} p{1.9em}<{\centering} p{1.9em}<{\centering} }
\zxline{1.3pt}
\multirow{2}*{Hand-crafted features} &\multicolumn{3}{c}{MMEW} &\multicolumn{3}{c}{\lyj{SAMM}} \\ \cline{2-7}
&KNN&SVM&ELM&\lyj{KNN}&\lyj{SVM}&\lyj{ELM}\\
\zxline{1pt}
\rowcolor{lightgray}LBP-TOP	\cite{51}	&\lyj{34.5}&\lyj{38.9}&\lyj{32.4}&\lyj{31.9}&\lyj{37.0}&\lyj{30.5}\\
DCP-TOP \cite{108}		&\lyj{37.6}&\lyj{42.5}&\lyj{36.2}&\lyj{36.1}&\lyj{36.8}&\lyj{32.8}\\
\rowcolor{lightgray} LHWP-TOP \cite{108} &\lyj{39.4} &\lyj{43.2}&\lyj{36.2}&\lyj{40.5}&\lyj{41.7}&\lyj{36.1}\\

RHWP-TOP \cite{108} &\lyj{37.6}&\lyj{45.9}&\lyj{37.8}&\lyj{36.1}&\lyj{38.1}&\lyj{43.8}\\
\rowcolor{lightgray} LBP-SIP \cite{57} &\lyj{35.3}&\lyj{43.9}&\lyj{36.6}&\lyj{36.5}&\lyj{37.4}&\lyj{35.8}\\
LBP-MOP \cite{95} &\lyj{43.9}&\lyj{41.5}&\lyj{34.6}&\lyj{32.4}&\lyj{35.3}&\lyj{34.7}\\
\rowcolor{lightgray} STLBP-IP \cite{82} &\lyj{36.6}&\lyj{46.3}&\lyj{46.6}&\lyj{35.7}&\lyj{42.9}&\lyj{39.3}\\
DiSTLBP-RIP \cite{103} &\lyj{39.0}&\lyj{44.0}&\lyj{41.5}&\lyj{42.9}&\lyj{46.2}&\lyj{42.9}\\
\rowcolor{lightgray} FDM \cite{97} &\lyj{30.8} &\lyj{34.6} &\lyj{31.4}&\lyj{33.3}&\lyj{34.1}&\lyj{32.4}\\
MDMO \cite{99}  &\lyj{53.4}&\lyj{60.6}&\lyj{65.7}&\lyj{44.1}&\lyj{50.0}&\lyj{50.0}\\
\rowcolor{lightgray} Sparse MDMO \cite{124} &\lyj{42.9} &\lyj{51.0} &\lyj{60.0}&\lyj{44.7}&\lyj{52.9}&\lyj{52.9}\\
\bottomrule
\zxline{1.3pt}
\end{tabular}
\vspace{-0.15in}
\label{tab:comparisons1}
\end{table}

\subsubsection{Comparisons of traditional methods}
\label{sssec:traditional-methods}

In this section, we evaluate the \lyj{recognition} performances of \lyj{representative} traditional methods that use \lyj{hand-crafted} features, \lyj{specifically} LBP-TOP \cite{51}, DCP-TOP \cite{108}, LHWP-TOP \cite{108}, RHWP-TOP \cite{108}, LBP-SIP \cite{57}, LBP-MOP \cite{95}, STLBP-IP \cite{82}, DiSTLBP-RIP \cite{103}, FDM \cite{97}, MDMO \cite{99} and Sparse MDMO \cite{124}. Meanwhile, we select KNN, SVM (with RBF kernel) and ELM \cite{58} as three representative classifiers, due to the fact that \lyj{these have been selected by most} previous micro-expression recognition works. All \lyj{of} the above-mentioned methods \lyj{follow their original settings outlined in the respective publications}, except for STLBP-IP and DiSTLBP-RIP. Based on \cite{82}, we divide each frame into $5\times5$ blocks before feature extraction using STLBP-IP.
In DiSTLBP-RIP, we produce the difference images of micro-expression image sequence, but we do not use robust principal component analysis (RPCA) \cite{RPCA} as in \cite{103}.

\begin{table}[t]
\footnotesize
\caption{\lyj{Recognition rates (\%) of micro-expressions using the state-of-the-art methods on MMEW and SAMM. The ranking of these methods are almost consistent on both datasets.}}
\centering
\renewcommand{\arraystretch}{1.2}
\begin{tabular}{l c c}
\zxline{1.3pt}
\multirow{2}*{Methods}&\multicolumn{2}{c}{Recognition rate (\%)} \\
\cline{2-3}&MMEW&SAMM\\
\zxline{1pt}
\rowcolor{lightgray} FDM \cite{97}	&\lyj{34.6}&\lyj{34.1}\\
ResNet10 \cite{ResNet10}	&\lyj{36.6}&\lyj{39.3}\\
\rowcolor{lightgray}Handcrafted features + deep features \cite{Handcrafted}	&\lyj{36.6}&\lyj{47.1}\\
LBP-TOP	\cite{51}	&\lyj{38.9}&\lyj{37.0}\\
\rowcolor{lightgray} Selective deep features \cite{100} &\lyj{39.0}&\lyj{42.9}\\
ELRCN \cite{115} 	&\lyj{41.5}&\lyj{46.2}\\
\rowcolor{lightgray}DCP-TOP \cite{108}		&\lyj{42.5}&\lyj{36.8}\\

ESCSTF \cite{ESCSTF} &\lyj{42.7}&\lyj{46.9}\\

\rowcolor{lightgray} LHWP-TOP \cite{108}	&\lyj{43.2}&\lyj{41.7}\\

LBP-MOP \cite{95}	&\lyj{43.9}&\lyj{35.3}\\

\rowcolor{lightgray} LBP-SIP \cite{57}	&\lyj{43.9}&\lyj{37.4}\\
DiSTLBP-RIP \cite{103}	&\lyj{44.0}&\lyj{46.2}\\
\rowcolor{lightgray}RHWP-TOP \cite{108}	&\lyj{45.9}&\lyj{38.1}\\
STLBP-IP \cite{82}	&\lyj{46.6}&\lyj{42.9}\\
\rowcolor{lightgray}ApexME \cite{ApexME}	&\lyj{48.8}&\lyj{50.0}\\

Transfer Learning \cite{Transfer} &\lyj{52.4}&\lyj{55.9}\\
\rowcolor{lightgray} Multi-task mid-level feature learning \cite{76}	&\lyj{54.2}&\lyj{55.0}\\
KGSL \cite{105}	&\lyj{56.9}&\lyj{48.6}\\
\rowcolor{lightgray} Sparse MDMO \cite{124} &\lyj{60.0}&\lyj{52.9}\\
MDMO \cite{99}		&\lyj{65.7}&\lyj{50.0}\\
\rowcolor{lightgray} DTSCNN \cite{112} &\lyj{65.9}&\lyj{69.2}\\
TLCNN \cite{114}	&\lyj{69.4}&\lyj{73.5}\\

\bottomrule
\zxline{1.3pt}
\end{tabular}
\vspace{-0.15in}
\label{tab:comparisons2}
\end{table}

Table \ref{tab:comparisons1} lists the best recognition rates of these hand-crafted features combined with different classifiers \lyj{on MMEW and SAMM}. \lyj{Except for LBP-TOP and sparse MDMO, the chosen parameters of each compared method are the same on MMEW and SAMM. The sizes of radii $R_x, R_y$ and $R_t$ on the three orthogonal planes (YT, XT and XY) of DCP-TOP, LHWP-TOP, RHWP-TOP, LBP-SIP, LBP-MOP are $(2,2,4)$, $(4,8,4)$, $(5,8,4)$, $(1,1,2)$, and $(1,1,2)$, respectively. The parameter for LBP-TOP is $(1,1,4)$ on both MMEW and SAMM. For STLBP-IP, the frame number of each video clip is set to 45, and the size of the linear mask of the one-dimensional local binary pattern (1DLBP) is set to 9. Moreover, for DiSTLBP-RIP, each frame is also divided into $5\times5$ blocks, and the dimension of the discriminative group-based feature using feature selection is 95. For FDM, each frame is divided into $6\times6$ blocks. Sparse MDMO achieves its best performance with the following parameters on MMEW (SAMM): the sparsity balance weight is 0.4 (0.4), the dictionary size is 256 (256), and the pooling parameter is 1 (0.1).}

\lyj{From Table \ref{tab:comparisons1}, it can be observed that generally speaking, better recognition accuracies are obtained by either SVM or ELM, but the performance of ELM is relatively unstable.
MDMO and Sparse MDMO have better recognition rates on both datasets. MDMO (65.7\%, 50\%) and sparse MDMO (60\%, 52.9\%) are the top two methods on both MMEW and SAMM. The reasons are as follows: MDMO and sparse MDMO are based on optical flow features extracted from ROI. Both methods utilize local statistical motion and spatial location information, and further apply the robust optical flow calculation method to capture the texture part of the image, followed by affine transformation. Therefore, the obtained optical flow field is insensitive to illumination conditions and head movement, which is important to have a good performance on SAMM.}

\subsubsection{Comparisons of state-of-the-art methods}
\label{sssec:deep-learning}

We \lyj{next} evaluate the \lyj{recognition} performance of state-of-the-art methods \lyj{(including deep learning methods, multi-task mid-level feature learning \cite{76} and KGSL \cite{105}) on MMEW and SAMM}. Table \ref{tab:comparisons2} summarizes the comparison results. \lyj{The handcrafted features in Table \ref{tab:comparisons1} are also kept in Table \ref{tab:comparisons2} for ease of comparison. We note that the ranking of these methods is almost consistent on both datasets.}

\lyj{For the multi-task mid-level feature learning \cite{76}}, LBP-TOP, LBP-MOP and an extension of LBP-MOP are used as the low-level features. \lyj{When} compared with the performance of the original low-level features (such as LBP-TOP \cite{51} and LBP-MOP \cite{95}), multi-task mid-level feature learning \cite{76} can enhance the discrimination ability of the original features \cite{51,95}. \lyj{In KGSL \cite{105},} hierarchical STLBP-IP is used as the spatio-temporal descriptor, which benefits from a hierarchical spatial division scheme, \lyj{so that} KGSL \cite{105} \lyj{obtains} the third best results among the non-deep-learning methods \lyj{(inferior to  MDMO and Sparse MDMO)}. \lyj{Moreover,} it is also noticeable that \lyj{all deep learning methods perform better than those utilizing handcrafted features}. \lyj{Setting up} deep learning methods \lyj{includes the} parameters, architectures, convolution layers, size of filters, batch normalization and so on, \lyj{the details of which are presented below.}

{\it Setting 1.} The learning rate of ResNet10 \cite{ResNet10} is 0.0001. The batch size is set to 8. ResNet10 includes 3 blocks and 2 fully connected layers; each block consists of two 3D convolution layers with convolution kernel of $3\times3\times3$ and one down-sampling layer with convolution kernel of $1\times1\times1$. The output sizes of the two fully connected layers are 128 and 6 respectively.

{\it Setting 2.} For handcrafted features + deep features \cite{Handcrafted}, we employ two scales (10, 20 pixels) and three orientations (0, 60 and 120 degrees), \lyj{resulting} in 6 different Gabor filters, and set the sizes of radii on the three orthogonal planes of LBP-TOP to (1, 1, 2). The deep CNN contains 5 convolutional layers and 3 fully connected layers. The convolution \lyj{kernel sizes} are $11\times11$, $5\times5$, $3\times3$, $3\times3$ and $3\times3$, \lyj{respectively}. The output sizes of the 3 fully connected layers are 9216, 4096 and 1000, respectively.

{\it Setting 3.}  ELRCN \cite{115} is based on VGG-16, which contains 13 convolutional layers ($3\times3$ conv) and 3 fully connected layers. The output sizes of 3 fully connected layers are 4096, 4096 and 2622, respectively. The output size of LSTM is 3000.

{\it Setting 4.} ApexME \cite{ApexME} is also based on VGG-16, with the same number of layers and convolution \lyj{kernel size} as ELRCN \cite{115}. The output sizes of the 3 fully connected layers are 256, 64 and 6, respectively. The batch size is set to 128. During the fine-tuning, the drop-out rate is set \lyj{to} 0.8 in order to avoid over-fitting.

{\it Setting 5.} In TLCNN \cite{114}, the CNN contains 5 convolutional layers ($11\times11$, $5\times5$, $3\times3$, $3\times3$, $3\times3$ conv) and 3 fully-connected layers. \lyj{Moreover,} there is a dropout layer after the first and second fully-connected layer. The learning rates for all training layers are all set to 0.001. The output sizes of the 3 fully connected layers are 256, 64 and 6, respectively.

{\it Setting 6.} In DTSCNN \cite{112}, the number of samples is extended to 90 for each class. All sequences are interpolated into different numbers of frames, i.e. 129 and 65. Their optical flow features are input into DTSCNN, which is a two-stream network \lyj{containing} 3D convolution and pooling units. The first network contains 4 convolutional layers ($3\times3\times8$, $3\times3\times3$, $3\times3\times3$, $3\times3\times4$ conv) and 1 fully connected layer, the output size \lyj{of which} is 6. Different from the first network, the convolution \lyj{kernel sizes} are $3\times3\times16$, $3\times3\times3$, $3\times3\times3$, and $3\times3\times4$.

{\it Setting 7.} For \lyj{selective} deep features \cite{100}, the CNN contains 10 convolutional layers ($3\times3\times3$ conv) and 2 fully-connected layers. After every two convolutional layers, there is a down-sampling layer ($1\times1\times1$ conv). In order to remove the irrelevant deep features trained by ImageNet and the facial expression dataset, evolutionary search is used to generalize the micro-expression data. The values \lyj{for} mutation probability, iteration number and population size are set to 0.02, 25 and 60, respectively.

{\it Setting 8.} The same ResNet10, LSTM and fully connected layers are used in transfer learning \cite{Transfer} and ESCSTF \cite{ESCSTF}. ResNet10 contains 3 blocks and 2 fully connected layers; \lyj{moreover,} each block consists of two 3D convolution layers ($3\times3\times3$ conv) and one down-sampling layer ($1\times1\times1$ conv). The output sizes are [8, 4, 256, 28, 28] respectively. The output of ResNet10 is fed to LSTM ($3\times3\times3$ conv). Finally, the output sizes of 2 fully connected layers are 1024 and 6. Although the two networks have the same structure, the training methods are different.

ResNet10 is pre-trained on ImageNet, then fine-tuned on macro- and micro-expression samples (16 interpolated frames for each sample) \lyj{through} transfer learning \cite{Transfer}. \lyj{Moreover,} in ESCSTF \cite{ESCSTF}, 5 key frames are used to train ResNet10 for each micro-expression sample; \lyj{namely,} the onset, onset to apex transition, apex, apex to offset transition and offset frames.

ResNet10 \cite{ResNet10}, TLCNN \cite{114}, transfer learning \cite{Transfer} and ESCSTF \cite{ESCSTF} use batch-normalization, which is \lyj{implemented before the} activation function (for example, ReLU). The purpose is to normalize the data to a mean of 0 and a variance of 1.

\begin{figure}[t]
\centering
\includegraphics[width=0.24\textwidth]{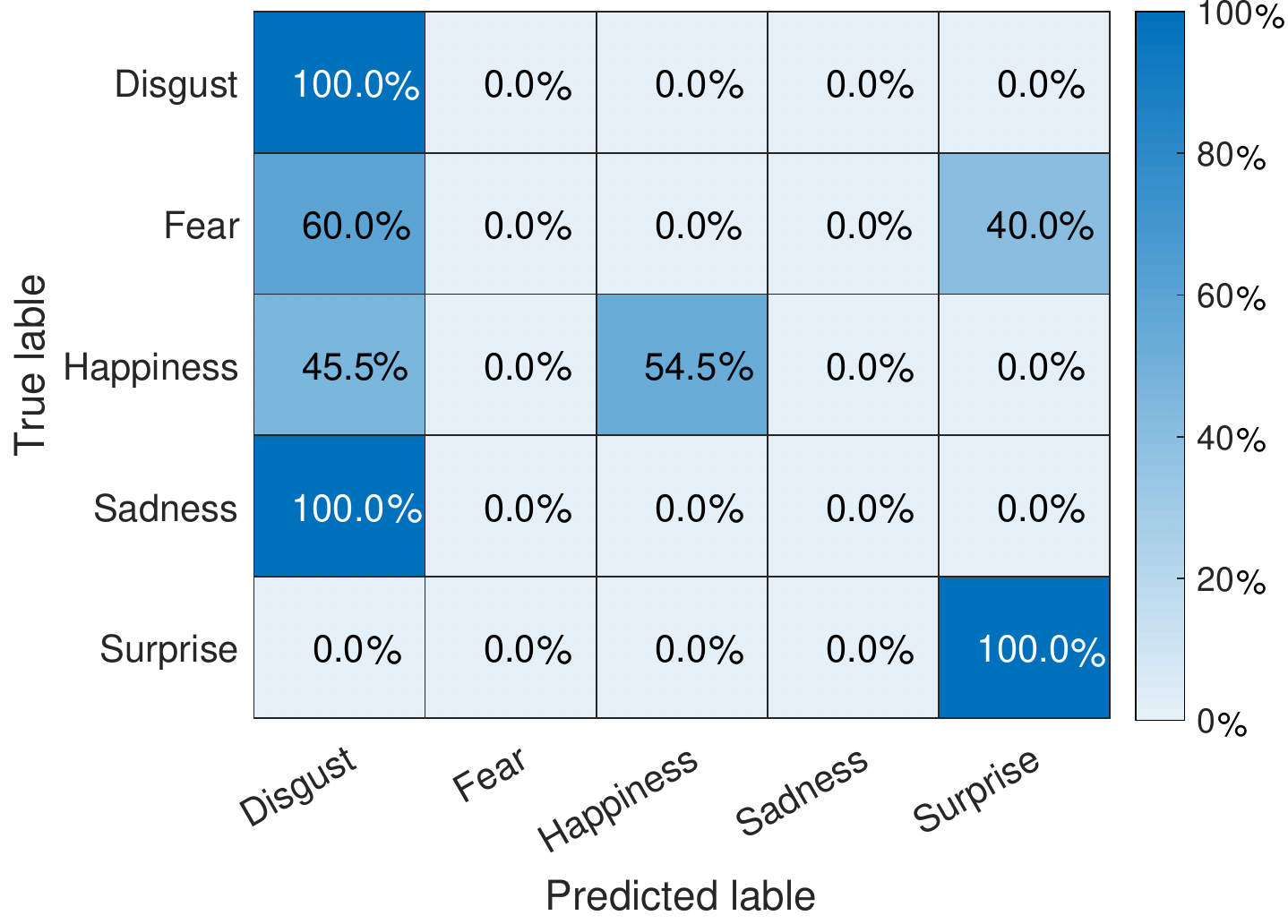}
\includegraphics[width=0.24\textwidth]{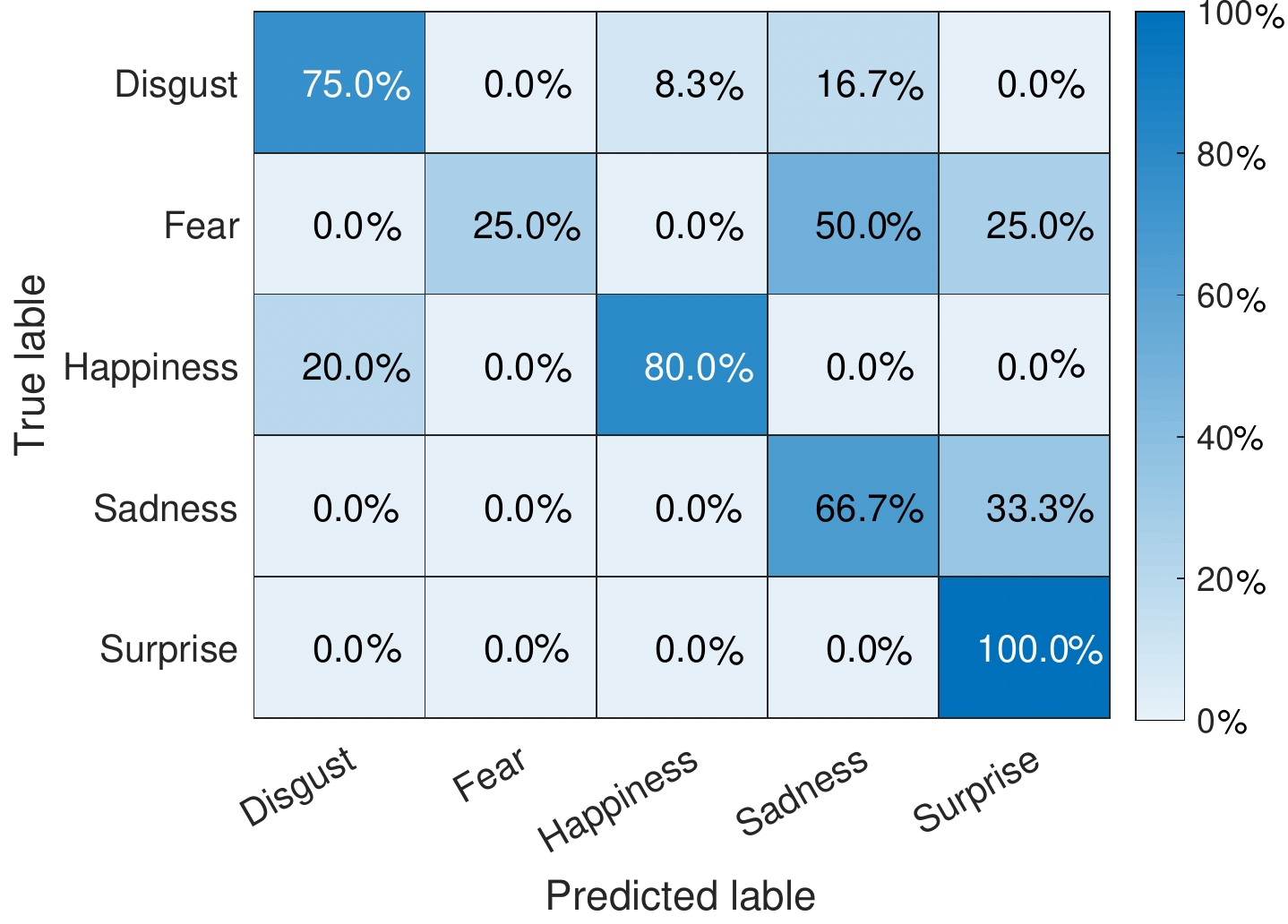}
\caption{Confusion matrices predicted by TLCNN on MMEW and SAMM. (a) MMEW; (b)SAMM.}
\vspace{-0.15in}
\label{fig:hunxiao}
\end{figure}

In Table \ref{tab:comparisons2}, it can be seen that TLCNN \cite{114} achieves the best \lyj{recognition} performance \lyj{(69.4$\%$ on MMEW and 73.5\% on SAMM)}. \lyj{This} is due to (1) \lyj{the use of} the macro-expression samples\footnote{\lyj{MMEW contains macro-expressions. For SAMM, the CK+ dataset \cite{CK+} are used as macro-expression samples.}} for pre-training, and (2) \lyj{the use of the} micro-expression samples fine-tuning, which solves the problem of the insufficient number of micro-expression samples. \lyj{Moreover}, LSTM is used to extract \lyj{the} discriminative dynamic characteristics from micro-expression sample sequences. In particular, we also present confusion matrices (see Figure \ref{fig:hunxiao}). We can see from Figure \ref{fig:hunxiao}(a) that all the ``disgust'' and ``surprise'' samples can be completely recognized on MMEW, instead the ``fear'' and ``sadness'' samples turn out to be harder samples to train. Because about four fifths of fear (16) and sadness (13) samples of MMEW were used for fine-tuning, where the number in brackets is the total sample number of each class, and the fine-tuning samples are too few.  A similar situation occurs in SAMM where the total sample number of fear and sadness are 7 and 3, respectively. The classes for instance ``fear'' and ``sadness'' are more likely to be inconsistent in the classification results (see Figure \ref{fig:hunxiao}(b)). Among deep learning methods, the baseline is ResNet10 \cite{ResNet10}, \lyj{which} can only obtain \lyj{a} recognition rate \lyj{of} \lyj{36.6\% on MMEW and 39.3\% on SAMM} because of over-fitting and \lyj{the} limited number of samples. The samples from the ImageNet dataset differ greatly from \lyj{the} micro-expression samples, and \lyj{a} limited fine-tuning effect \lyj{spoils} the recognition performance; therefore, transfer learning \cite{Transfer} achieves recognition accuracies of 52.4\% on MMEW \lyj{and 55.9\% on SAMM}.

\subsection{Encoding micro- from macro-expressions}
\label{ssec:evaluation-cross}

The success of TLCNN \cite{114} demonstrates that the macro-expressions are relevant for pre-training the network; this effectively alleviates the famous 3S (small sample size) problem inherent in micro-expression datasets. We also performed three group experiments: (1) CK+ was used for pre-training while MMEW (Micro) was applied for fine-tuning and testing, noted as ``CK+ $\rightarrow$ MMEW (Micro)''; (2) MMEW (Macro) was used for pre-training while MMEW (Micro) was applied for fine-tuning and testing, noted as ``MMEW (Macro) $\rightarrow$ MMEW (Micro)''; (3) CK+ was used for pre-training while SAMM was applied for fine-tuning and testing, noted as ``CK+ $\rightarrow$ SAMM''. The training and testing datasets of MMEW dataset were set as above-mentioned. Table \ref{tab:Data} lists the data source and accuracy of the pre-training, fine-tuning and testing phases for each experiment. MMEW (Macro) $\rightarrow$ MMEW (Micro) $\textgreater$ CK+ $\rightarrow$ MMEW (Micro) (with descending effect) indicate that the knowledge of macro-expressions is effective for micro-expression recognition, and using macro- and micro-expressions from the same dataset to pre-trained and fine-tune has better performance than using those from different datasets. This also demonstrates our original intention to establish MMEW, i.e. dataset containing both macro- and micro-expression of the same subject is required in order to transfer macro-expression knowledge to assist micro-expression recognition. The advantage of the new released MMEW dataset brings up an interesting problem:

\begin{table*}[t]
\footnotesize
\centering
\caption{Data sources and accuracies of the pre-training, fine-tuning and testing phases.}
\renewcommand{\arraystretch}{1.2}
\begin{tabular}{p{16.5em} p{8em} p{4.5em} p{8em} p{4.5em} p{8em} p{4.5em}}
\zxline{1.3pt}
\multirow{2}*{Experiments}&\multicolumn{2}{c}{Pre-training} &\multicolumn{2}{c}{Fine-tuning}&\multicolumn{2}{c}{Testing}\\
\cline{2-7}&Data source&Acc.&Data source&Acc.&Data source&Acc.\\
\zxline{1pt}
\rowcolor{lightgray} CK+ $\rightarrow$ MMEW (Micro) &CK+ &99.9\% &MMEW (Micro)&99.8\%&MMEW (Micro)&65.6\% \\
MMEW (Macro) $\rightarrow$ MMEW (Micro)	&MMEW (Macro)&92.0\%&MMEW (Micro)&96.6\%&MMEW (Micro)&69.4\%\\
\rowcolor{lightgray}CK+ $\rightarrow$ SAMM &CK+ &99.0\%&SAMM  &99.7\%&SAMM&73.5\% \\
\bottomrule
\zxline{1.3pt}
\end{tabular}
\vspace{-0.15in}
\label{tab:Data}
\end{table*}

{\it ``Do the macro-expressions of the same person help with recognizing his/her own micro-expressions?''}
Since MMEW contains the macro- and micro-expressions of the same participants, we also study this problem in a subject-dependent way. Note that although CAS(ME)$^2$ also contains the macro- and micro-expressions of the same participants, the number of micro-expression samples is small (only 57), so that CAS(ME)$^2$ is not suitable for studying this problem.

Accordingly, by utilizing MMEW, we conducted a preliminary experiment to evaluate different combinations of macro- and micro-expressions. (1) {\it Subject-independent evaluation:} The final average recognition rate was 69.4\%.  (2) {\it Subject-dependent evaluation:} All the micro-expression samples in MMEW were randomly divided into five fold, and for each fold, all macro-expressions in MMEW were used for pre-training. Then five-fold cross validation was also used for performance evaluation. The final average recognition rate was increased to 87.2\%. These results demonstrate that macro-expressions of the same person are more relevant than macro-expressions of different persons for pre-training the network. Therefore, MMEW can be used to explore new research directions, including subject-independent and subject-dependent encoding from macro- to micro-expressions.

\section{Future directions}
\label{sec:direction}

Despite \lyj{the significant progress made in} micro-expression analysis over the last decade, several outstanding issues and new avenues exist for future development. Below, we propose \lyj{some potential} research directions.


\lyj{
{\it Privacy-protection micro-expression analysis.}
Akin to macro-expressions, micro-expressions also constitute a kind of private facial information. By providing adequate privacy and security, micro-expression spotting and recognition conducted by utilizing federated learning from decentralized data distributed across private user devices deserves study in its own right \cite{Yang2019}.
}

{\it Utilizability of macro-expressions for micro-expression analysis.}
Both macro- and micro-expressions can be characterized according to the emotional facial action coding system. Considering the benefits offered by the new MMEW dataset, it would be interesting to explore the mutual effect of macro- and micro-expressions, particularly those from the same subject. Furthermore, as previous research has indicated that many deceptive behaviors are dependent on individual differences \cite{Warren2009Detecting}, we conjecture that micro-expression behavior is likely subject-dependent and MMEW dataset provides a new platform for both subject-dependent and subject-independent research in the future.

\lyj{{\it Standardized datasets.}}
There are two major problems in the current micro-expression datasets. First, \lyj{the existing number of micro-expression samples is too limited to facilitate proper training,} since inducing micro-expressions is \lyj{quite} difficult. Researchers often require participants to watch emotional videos to \lyj{elicit} their emotions, and even to disguise their expressions. However, some participants may not \lyj{exhibit micro-expressions under these circumstances, or may only exhibit them rarely}. In addition, the encoding/labeling of micro-expressions is time-consuming and laborious, since it requires the viewer \lyj{to} (1) watch the video at a slow speed and (2) select the onset, apex and offset frames of the facial motion, then calculate their duration. Consequently, there is no uniform standard \lyj{available} to annotate the emotion of micro-expressions.
Second, \lyj{owing to the poor quality of the videos in many existing micro-expression datasets, the full details of the low-intensity micro-expressions cannot be fully captured}. Therefore, videos with higher temporal and spatial resolutions are needed for \lyj{future algorithm design}.

\lyj{{\it Data augmentation.}} Some data augmentation tricks \lyj{used by the} deep learning community \lyj{could be} helpful to increase the \lyj{amount of available} micro-expression data. For example, image rotation, translating, cropping and similar techniques will not change the labels of micro-expression but will increase the amount of available micro-expression data, leading to a potential performance improvement.

\lyj{{\it GAN-based sample generation.}
Deep learning methods achieve superior performance in facial expression recognition tasks. For example, Deng et al. \cite{li2018deep} present an extensive and detailed review of state-of-the-art deep learning techniques for facial expression recognition. However, it is difficult to apply them to micro-expression recognition, since they will suffer due to the lack of sufficient training samples. In addition to manipulating images for data augmentation, it would be feasible to address this issue through utilizing generative adversarial networks (GANs) in order to generate a large number of pseudo-micro-expression samples, provided that we can define some criteria to ensure that the generated samples are indeed micro-expressions.}

\lyj{{\it Multi-task learning.} The extraction of micro-expression heavily depends on the ability to detect facial feature points that occupy characteristic positions, the semantic location of which has been predefined. The reason is that the motion amplitude of micro-expression is quite subtle, and facial feature points detection can reduce the effect caused by head movement during data preprocessing. It would therefore be useful in the future to design an end-to-end model capable of both learning the motion amplitude of micro-expression and detecting facial feature points. In this way, we could potentially alleviate the cost of micro-expression annotations.}

\lyj{
{\it Explainable micro-expression analysis.} Although deep learning methods have received increasing attention and achieved good performance in micro-expression analysis, these deep models are usually treated as black boxes and have poor interpretability and explainability. In the critical applications such as lie detection and criminal justice, explainability is very important to help human understand the reasons behind predictions \cite{Rudin2019}.
}

\section{Conclusion}
\label{sec:conclusion}

Micro-expression analysis \lyj{has} a wide range of \lyj{potential real-world} applications; \lyj{for example,} enabling people to detect micro-expressions in daily life and \lyj{develop} a good interpretation/understanding of what \lyj{lies} behind micro-expressions. \lyj{To make micro-expression analysis useful in practice,} we need to develop robust algorithms with valid and reliable samples in order to make the spotting and recognition of micro-expressions applicable to real situations. \lyj{Accordingly}, in this survey, we review the current research on spontaneous facial micro-expression analysis (including \lyj{datasets,} features and algorithms) and propose a new dataset, \lyj{MMEW,} for micro-expression recognition. We \lyj{further} compare the performance of \lyj{existing} state-of-the-art methods, analyze the potential, and highlight the outstanding issues \lyj{for future research on} micro-expression analysis. Micro-expression analysis has \lyj{recently} become an active research area; \lyj{accordingly,} we hope this survey can help researchers, \lyj{as a starting point, to review the developments in the state-of-the-art} and identify possible directions for \lyj{their} future research.

\ifCLASSOPTIONcaptionsoff
  \newpage
\fi


\bibliographystyle{IEEEtran}
\bibliography{ref}

\begin{IEEEbiography}[{\includegraphics[width=1in,height=1.25in,clip,keepaspectratio]{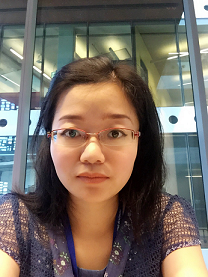}}]{Xianye Ben}
received the Ph.D. degree from the College of Automation, Harbin Engineering University, in 2010. She is now a Professor in the School of Information Science and Engineering, Shandong University, China. She has published more than 90 papers in major journals and conferences, such as IEEE T-IP, IEEE T-CSVT, IEEE T-MM, PR, CVPR, etc.
She received the Excellent Doctorial Dissertation awarded by Harbin Engineering University. She was also enrolled by the Young Scholars Program of Shandong University.
\end{IEEEbiography}

\begin{IEEEbiography}[{\includegraphics[width=1in,height=1.25in,clip,keepaspectratio]{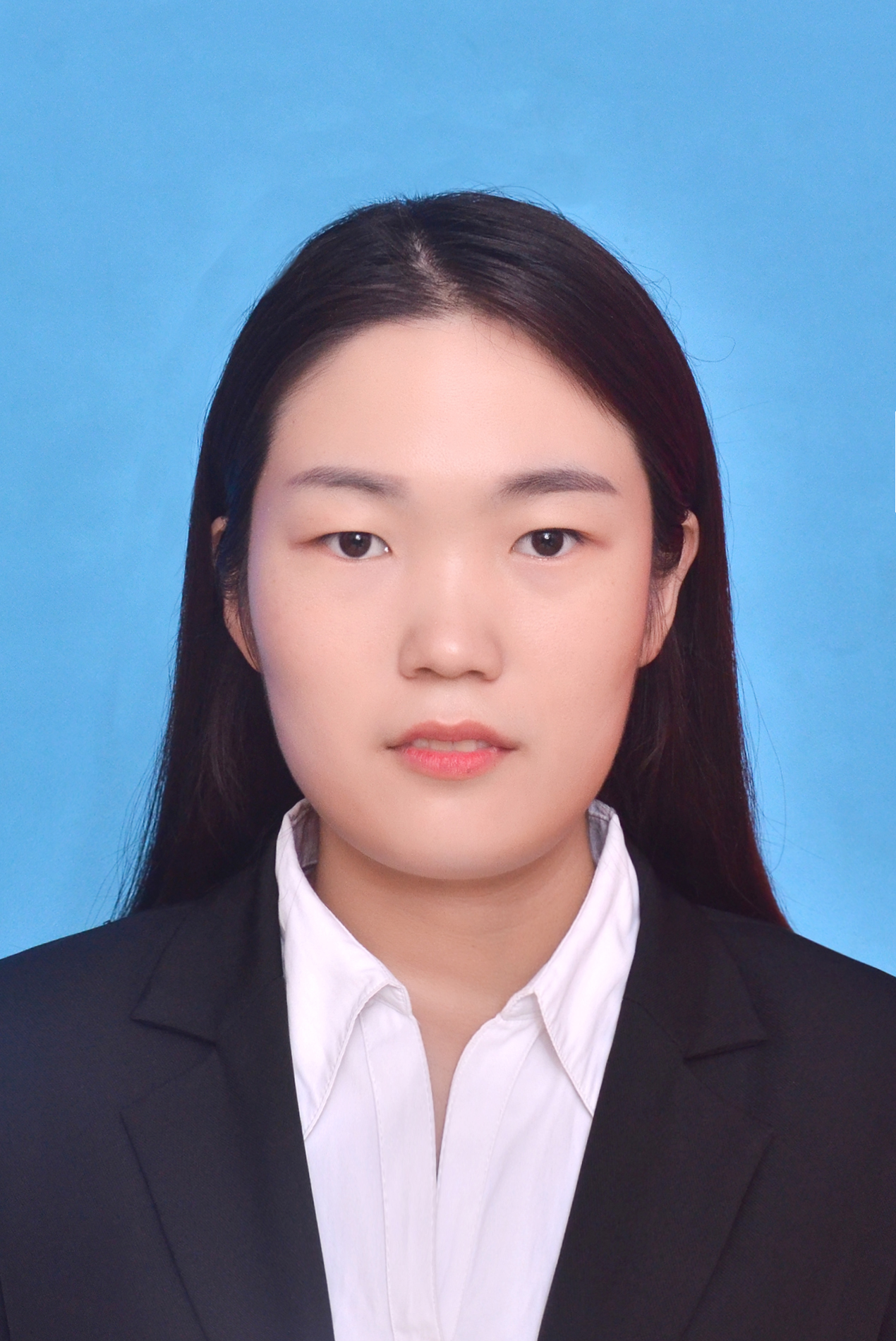}}]{Yi Ren}
received the B.S. degree in electronic and information engineering from the School of Physics and Electronic Science, Shandong Normal University, Jinan, China, in 2016. She is a third-year graduate students majored in Information and Communication Engineering in the School of Information Science and Engineering, Shandong University, Qingdao, China. Her current research interests include micro-expression spotting/detection and recognition.
\end{IEEEbiography}
\vspace{-0.25in}

\begin{IEEEbiography}[{\includegraphics[width=1in,height=1.25in,clip,keepaspectratio]{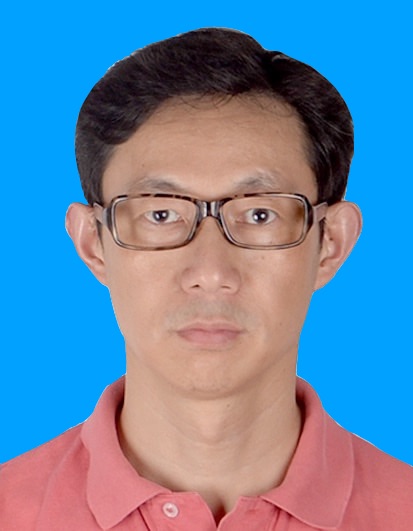}}]{Junping Zhang}
(M'05) is a professor at School of Computer Science, Fudan University since 2011. His research interests include machine learning, image processing, biometric authentication, and intelligent transportation systems. He has been an associate editor of the IEEE INTELLIGENT SYSTEMS since 2009.  He has widely published in highly ranked international journals such as IEEE TPAMI and IEEE TNNLS, and leading international conferences such as ICML, AAAI and ECCV.
\end{IEEEbiography}
\vspace{-0.25in}

\begin{IEEEbiography}[{\includegraphics[width=1in,height=1.25in,clip,keepaspectratio]{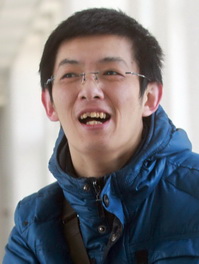}}]{Su-Jing Wang}
(SM'19) received the Ph.D. degree from the College of Computer Science and Technology of Jilin University in 2012. He is now an Associate Researcher in Institute of Psychology, Chinese Academy of Sciences. He was called as \emph{Chinese Hawkin} by the Xinhua News Agency. His current research interests include pattern recognition, computer vision and machine learning. He serves as an associate editor of Neurocomputing (Elsevier).
\end{IEEEbiography}
\vspace{-0.25in}

\begin{IEEEbiography}[{\includegraphics[width=1in,height=1.25in,clip,keepaspectratio]{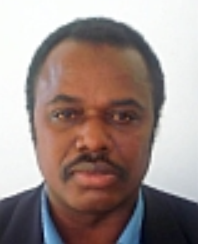}}]{Kidiyo Kpalma}
was born in Togo in 1962. He joined the Institut National des Sciences Appliquees de Rennes (INSA), and received the PhD degree in 1992. In 1994, he became Associate Professor at INSA where he teaches Analog/Digital Signal processing an Automatic. After his HDR (Habilitation a diriger des recherches) degree from the University of Rennes 1 in 2009, he obtain the position of Professor with INSA since 2014. As a member of IETR UMR CNRS 6164, his research interests include pattern recognition, semantic image segmentation, facial micro-expression and saliency object detection.
\end{IEEEbiography}
\vspace{-0.25in}

\begin{IEEEbiography}[{\includegraphics[width=1in,height=1.25in,clip,keepaspectratio]{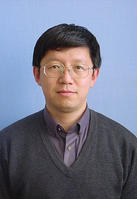}}]{Weixiao Meng}
(SM'10) received the B.Eng., M. Eng., and Ph.D. degrees from Harbin Institute of Technology (HIT), Harbin, China, in 1990, 1995, and 2000, respectively. He is a professor of the School of Electronics and Information Engineering of HIT. He is the Chair of IEEE Communications Society Harbin Chapter and a Fellow of the China Institute of Electronics. He won Chapter of the Year Award, Asia Pacific Region Chapter Achievement Award and Member \& Global Activities Contribution Award in 2018.
\end{IEEEbiography}
\vspace{-0.25in}

\begin{IEEEbiography}[{\includegraphics[width=1in,height=1.25in,clip,keepaspectratio]{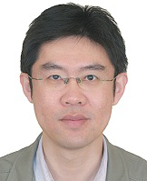}}]{Yong-Jin Liu}
(SM'16) is a tenured full professor with Department of Computer Science and Technology, Tsinghua University, China.
He received his B.Eng degree from Tianjin University, China, in 1998, and the PhD degree from the Hong Kong University of Science and Technology, Hong Kong, China, in 2004.
His research interests include cognition computation, computational geometry, computer graphics and computer vision.
\end{IEEEbiography}

\end{document}